\pdfoutput=1
\documentclass[manuscript,screen,nonacm]{acmart}

\setcopyright{none}
\usepackage{booktabs}
\usepackage{tabularx}
\usepackage{xcolor}
\usepackage{tikz}
\usetikzlibrary{arrows.meta,positioning,fit,calc}
\usepackage{enumitem}
\setlist{nosep,leftmargin=1.4em}

\newcommand{\E}{\mathbb{E}}
\newif\ifshowchanges \showchangesfalse
\newcommand{\chg}[1]{{\ifshowchanges\color{blue!75!black}\fi#1}}
\newcommand{\chgr}[1]{{\ifshowchanges\color{red!75!black}\fi#1}}
\newif\ifshowcr \showcrfalse
\DeclareRobustCommand{\crx}[1]{{\ifshowcr\color{magenta!75!black}\fi#1}}

\newcommand{\phead}[1]{\textbf{#1}}
\makeatletter
\gdef\cbd@authsep{}
\def\@typeset@author@line{%
  \andify\@currentauthors
  \ifx\@currentauthors\@empty\else
    \cbd@authsep\@currentauthors\gdef\cbd@authsep{\unskip; }%
  \fi
  \def\@currentauthors{}%
  \ifx\@currentaffiliations\@empty\else
    \andify\@currentaffiliations
      \unskip, {\@currentaffiliations}%
  \fi
  \def\@currentaffiliations{}}
\makeatother
\begin{document}

\title[Local Audits in Outcome-Learned Pockets]{Where Is a Prediction Model Miscalibrated, and for Whom? Local Audits in Outcome-Learned Pockets}

\author{Bernardo Modenesi}
\email{bernardo.modenesi@utah.edu}
\affiliation{\institution{University of Utah}\city{Salt Lake City}\state{UT}\country{USA}}
\author{Taoyi Chen}
\email{chen.15500@osu.edu}
\affiliation{\institution{Ohio State University}\city{Columbus}\state{OH}\country{USA}}
\author{Sima Najafzadehkhoei}
\email{sima.njf@utah.edu}
\affiliation{\institution{University of Utah}\city{Salt Lake City}\state{UT}\country{USA}}
\author{Lucia Wang}
\email{wang.luciay@gmail.com}
\affiliation{\institution{rpotential.ai}\country{USA}}
\author{Ameya Diwan}
\email{ameyakdiwan@gmail.com}
\affiliation{\institution{Google}\country{USA}}
\author{Beatrice Knudsen}
\email{beatrice.knudsen@path.utah.edu}
\affiliation{\institution{University of Utah}\city{Salt Lake City}\state{UT}\country{USA}}

\begin{abstract}
Overall and group-level averages, and ranking metrics such as the AUC, can hide sets of observations for which a
prediction model systematically over- or under-predicts, because concentrated errors barely move an average and
opposite errors offset each other. We present an audit that asks where a model is miscalibrated (in which sets of similar observations it
over- or under-predicts on average) and for whom (whether its average prediction error differs between
demographic groups, such as by race or sex, within the same set). It groups observations into \emph{pockets}, sets
that are similar in how the outcome is predicted from their features, using random-forest similarity, which \crx{in effect} weighs
each feature by its importance for predicting the outcome instead of relying on hand-chosen cuts of the data\crx{. A}
second set of pockets, built on features weighted by ridge-regression coefficients, catches errors that follow one
strong predictor, and a combined audit uses both. Pockets are learned on one sample and
tested on another, with confidence intervals that hold jointly across pockets and allow for correlation within units
such as hospitals. For each pocket, the audit reports the average prediction error, its gaps between demographic
groups, the error sizes ruled out, and whether the data suffice to test. It \crx{works from} features, outcomes and fixed
predictions, not model internals, which can be proprietary or undisclosed. In 17 simulation scenarios (continuous and binary
outcomes, nonlinear risk, missing values, dependent observations, real covariates) with an error planted for one
demographic group (\emph{for whom}), the combined audit located it better than the strongest alternative grouping in
15, with false-alarm rates near or below 5\%. In 15 scenarios with an error planted for everyone in one feature region
(\emph{where}), it localized the error better than every available alternative in 12 (at the smaller of two
error sizes), including errors that offset overall; it lost to a grouping learned from one model's residuals when the
error was unrelated to the outcome, and with few units of correlated observations it needed a stricter threshold. We
then audit two deployed scores with documented problems, asking what a local audit adds. First, in public synthetic
data from the care-management algorithm studied by Obermeyer et al., with its cost-based score mapped to a prediction
of chronic illness, we found no detectable overall error but under-prediction of illness in the sickest pockets and
over-prediction in the others. The pockets distinguished Black patients' concentration where the score
under-predicted for both groups from a Black--White gap within pockets. Second, an exploratory analysis located
APACHE~IVa's previously reported over-prediction of \crx{hospital mortality for ICU stays} in a
low-risk pocket holding half of the \crx{audit} stays with an APACHE prediction, where predicted mortality was more than twice the
observed rate\crx{, and less firmly in a pocket of mostly ventilated stays}; none was detected in a moderate-risk pocket. Two recalibrated chest X-ray classifiers also showed
offsetting local errors behind no detectable overall error. \crx{A forest trained on the}
outcome, not on any model's errors, needs no rescaling or imputation of numeric inputs, gives many models one
frozen map, and \crx{lets groups be compared} within shared pockets.

\end{abstract}



\maketitle
\renewcommand{\shortauthors}{Modenesi et al.}

\section{Introduction}
\label{sec:intro}

An AI model can be accurate on average yet systematically wrong for a particular kind of observation. \crx{Such errors matter when a predicted value guides a decision.} For example, health systems use commercial algorithms to choose which
patients enter care-management programs; such tools were estimated in 2019 to be applied to about 200 million people
a year in the United States \cite{obermeyer2019dissecting}. Intensive care units benchmark mortality with severity
scores such as APACHE~IVa \cite{zimmerman2006apache}. When a model reports a 20\% risk, about one in five such cases should
have the event; a model with this property is \emph{calibrated}, and a miscalibrated model systematically over- or
under-predicts. A useful audit must locate where this happens, show whose predictions are affected, and distinguish a
small error from too little data to tell.

Calibration is usually checked overall or within a few predefined groups, alongside ranking metrics such
as the AUC. For \emph{where}, these checks can miss unrecognized, poorly served subsets, known in medical
imaging as \emph{hidden stratification} \cite{oakdenrayner2020hidden}, and errors of opposite sign that offset each
other in any average. Slice-discovery methods search for such subsets, but several target one model's errors, so the
subsets found can depend on the model audited (Section~\ref{sec:related}). For
\emph{whom}, audits have moved from predefined groups to their intersections \cite{buolamwini2018gender} and to large
classes of subgroups defined by combined attributes \cite{kearns2018gerrymandering,cherian2024fairaudit}, and ICU severity scores were found to
over-predict mortality for Black and Hispanic patients \cite{sarkar2021apache}. Yet a gap between groups can arise
in two ways: one group is concentrated where the model errs for everyone, or the model errs differently for members
of the two groups who are alike in what predicts the outcome. The two call for different responses, and a
group-level comparison cannot tell them apart. The algorithm studied by \citet{obermeyer2019dissecting} shows what is at stake: its score was approximately
calibrated for costs across racial groups, but because Black patients incurred lower costs at the same level of
illness, it understated their health needs, and an audit against costs alone could not see it. In public synthetic
data from that study, with the score mapped to a prediction of illness, no overall error is detectable, yet our audit
finds under-prediction \crx{in a pocket holding} a third of patient-years and over-prediction elsewhere; Black patients are
over-represented in that \crx{pocket}, and a gap remains within it (Section~\ref{sec:apps}).

We ask two questions of a model's predictions within \emph{pockets}, sets of observations similar in how their
features predict the outcome. \emph{Where} is the model miscalibrated? And \emph{for whom}: within a pocket, does the
average signed prediction error differ between groups defined by a recorded attribute, such as race or sex? Following the literature
on differential item functioning \cite{holland1988differential}, we call the group under study the \emph{focal}
group and the group it is compared with the \emph{reference} group. Answering both poses three practical challenges:
(1) \emph{Where should we look?} Checking every combination of features is intractable. (2) \emph{Is there enough
data?} A subpopulation with few observations, few positive outcomes or few focal-group members cannot support a
comparison, and pretending otherwise produces false alarms. (3) \emph{How sure are we?} Auditing many subpopulations
multiplies the chances of a false alarm, and observations from the same unit, such as repeated records of one person
or patients of one hospital, are not independent.

Our audit learns the pockets from the outcome on a \emph{discovery} sample used for nothing else, and tests them
on a separate \emph{audit} sample. It estimates each pocket's \emph{local calibration error} $\mu$, the average
prediction error (predicted minus observed outcome), and \emph{local calibration gap} $\delta$, the difference in that
error between the focal and reference groups, with confidence intervals that hold simultaneously across pockets and
allow for correlated observations within units; an estimate is \emph{flagged} when its interval excludes zero.
Keeping the samples apart costs data, but no pocket is tested on the outcomes that suggested it; whether the
intervals then hold at their stated level is checked by simulation.

The main grouping comes from a random forest trained to predict the outcome: two observations are close when many
trees send them to the same leaf \cite{breiman2001rf}. Because trees split mostly on features that improve the
prediction, observations that differ only in irrelevant features \crx{still tend to share leaves}, so this similarity \crx{in effect} weighs
each feature by its relevance to the outcome (a built-in feature selection), whereas K-means on standardized features
weighs every feature equally. It assumes no distance or distribution for the features, needs no rescaling or imputation of
numeric features and scales to large data through a sparse nearest-neighbor graph. Because it spreads over all outcome-relevant features, it can blur a failure that follows one
strong predictor, so a simpler \emph{linear} grouping clusters observations on features weighted by their
coefficients in a ridge regression of the outcome, and our default, the \emph{combined audit}, runs both. Neither grouping uses
the group attribute or any model's errors as an input; group counts and predictions enter only the \emph{support}
check, which decides whether a pocket has enough data to test. The same pockets can therefore audit several models from their predictions alone, even when a model's internals
are proprietary or undisclosed. Learning pockets from the outcome also makes the answers interpretable: a pocket's mean error
checks calibration among cases grouped by what predicts the outcome, and a gap within it compares groups that share
the pocket.

Our contributions are:
\begin{itemize}
  \item \textbf{Reusable outcome-learned audit maps.} A discovery-only procedure learns pockets from the outcome
        and freezes the rule that assigns observations to them, so several models can be compared, from their
        predictions alone, through each pocket's mean prediction error and its between-group gap
        (Sections~\ref{sec:method} and~\ref{sec:inference}).
  \item \textbf{Statistical evidence, not only flags.} Simultaneous intervals that allow for correlated
        observations within units such as hospitals; when a pocket's simultaneous interval lies within $(-t, t)$, a statement that pocket-average errors of size
        $t$ or larger are ruled out; and an explicit report of pockets too thin to test.
  \item \textbf{An evaluation that reports its losses} (Section~\ref{sec:sims}). The combined audit located gaps planted
        between demographic groups better than the strongest alternative grouping in 15 of 17 scenarios, and local
        miscalibration better than every available alternative in 12 of 15 at the smaller of two error sizes. It lost to groupings learned from the
        model's own errors when the error sat on an outcome-irrelevant feature or in one tail of a U-shaped feature.
        False-flag rates were near or below 5\%, except for $\mu$ under strong within-unit dependence, which needs a
        stricter level.
  \item \textbf{Three applications} (Section~\ref{sec:apps}), two of them deployed scores with documented problems,
        where we ask what a local audit adds. Synthetic care-management data show
        offsetting local errors behind a reassuring overall check, and separate Black patients' concentration in an
        under-predicted pocket from the gap within pockets; an exploratory
        ICU audit locates APACHE~IVa's previously reported over-prediction in two of three testable pockets, under hospital
        dependence and limited support; a briefer audit finds offsetting local errors in
        two chest X-ray classifiers.
\end{itemize}

\section{Related Work}
\label{sec:related}

\phead{Calibration and locating model failures.} In the calibration hierarchy of
\citet{vancalster2016hierarchy}, our local error is \emph{mean} calibration within a pocket, weaker than
calibration at every predicted value, so we treat it as a screening quantity. Slice-discovery methods search for
regions of hidden stratification \cite{oakdenrayner2020hidden} with interpretable feature slices and significance tests
\cite{chung2020slicefinder,sagadeeva2021sliceline} or with clusters of learned representations
\cite{sohoni2020george,eyuboglu2022domino,deon2022spotlight}; several organize regions around one model's errors.
LOGAN clusters observations to expose local group gaps \cite{zhao2020logan}, and forest-guided clustering groups
observations through random-forest decision paths \cite{fgc2025}. Our pockets are organized around
outcome-relevant similarity and frozen before audit outcomes are read, so the same pockets can compare several
models; the price is that they can miss errors unrelated to the outcome.
Multicalibration asks for calibration on every subpopulation a function class can identify
\cite{hebertjohnson2018multicalibration}, and multiaccuracy for a zero mean error on each
\cite{kim2019multiaccuracy}. Our target is narrower, a finite set of learned pockets, for which we report held-out
uncertainty and explicit limits on what the data can test.

\phead{Fairness and audit inference.} Calibration within groups cannot in general be combined with equal
error rates across groups unless base rates are equal or prediction is perfect
\cite{kleinberg2017inherent,chouldechova2017fair,pleiss2017fairness}, so an audit must choose what it measures; we
audit calibration because decisions act on the predicted value itself
\cite{vancalster2019calibration,pfohl2021empirical}. Subgroup audits seek fairness violations over
structured classes of protected-attribute subgroups \cite{kearns2018gerrymandering}, and
\citet{cherian2024fairaudit} treat auditing as inference with simultaneous bootstrap bounds over rich subgroup
classes. We distinguish pocket composition from within-pocket gaps, comparing group-average errors in
outcome-learned pockets fixed before audit outcomes are read. We use the same max-statistic logic
\cite{westfall1993resampling} on a finite set of pockets learned on independent data, as honest trees separate the samples
that form and estimate subgroups \cite{athey2016recursive,wager2018estimation}, and calibrate it for correlated
observations with a wild cluster bootstrap \cite{cameron2008bootstrap,mackinnon2017wild,mackinnon2023cluster}.

\phead{Supervised similarity.} Random forests act as adaptive nearest-neighbor methods whose neighborhoods
stretch along directions irrelevant to the outcome and shrink along relevant ones
\cite{breiman2001rf,lin2006random}. Individual
fairness presupposes a task-specific similarity measure \cite{dwork2012fairness,ilvento2020metric}; we use forest
similarity only to form pockets, with no guarantee for individuals.

\section{Method}
\label{sec:method}

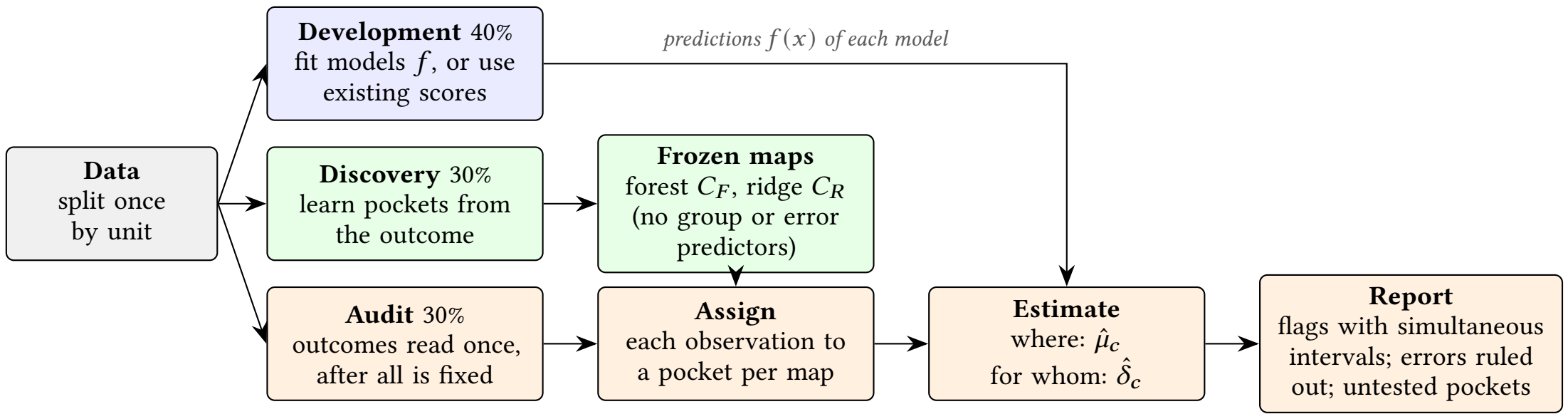
\begin{figure}[t]
\centering
\begin{tikzpicture}[
  font=\footnotesize, >={Stealth[length=2mm]},
  box/.style={draw, rounded corners=2pt, align=center, minimum height=1.05cm, text width=2.35cm, fill=#1},
  box/.default=white, lab/.style={font=\scriptsize\itshape, text=black!65}]
  \node[box=gray!12, text width=1.75cm] (data) {\textbf{Data}\\split once\\by unit};
  \node[box=blue!8, right=0.45cm of data, yshift=1.3cm] (dev) {\textbf{Development} 40\%\\fit models $f$, or use\\existing scores};
  \node[box=green!10, right=0.45cm of data] (disc) {\textbf{Discovery} 30\%\\learn pockets from\\the outcome};
  \node[box=orange!12, right=0.45cm of data, yshift=-1.3cm] (aud) {\textbf{Audit} 30\%\\outcomes read once,\\after all is fixed};
  \draw[->] (data.east) -- (dev.west);
  \draw[->] (data.east) -- (disc.west);
  \draw[->] (data.east) -- (aud.west);
  \node[box=green!10, right=0.5cm of disc] (freeze) {\textbf{Frozen maps}\\forest $C_F$, ridge $C_R$\\(no group or error\\predictors)};
  \draw[->] (disc) -- (freeze);
  \node[box=orange!12, right=0.5cm of aud] (assign) {\textbf{Assign}\\each observation to\\a pocket per map};
  \draw[->] (aud) -- (assign);
  \draw[->] (freeze.south) -- (assign.north);
  \node[box=orange!12, right=0.5cm of assign] (est) {\textbf{Estimate}\\where: $\hat\mu_c$\\for whom: $\hat\delta_c$};
  \draw[->] (assign) -- (est);
  \node[box=orange!12, right=0.5cm of est, text width=2.6cm] (inf) {\textbf{Report}\\flags with simultaneous\\intervals; errors ruled\\out; untested pockets};
  \draw[->] (est) -- (inf);
  \draw[->] (dev.east) -| node[lab, pos=.25, above] {predictions $f(x)$ of each model} (est.north);
\end{tikzpicture}
\caption{The audit at a glance. Whole units (e.g., hospitals) are split once into development, discovery and audit
roles. Two pocket maps are learned from the outcome on discovery data, without the group attribute or any model's
errors as inputs (group counts and predictions enter only the support check), and frozen: forest pockets $C_F$ and ridge pockets $C_R$. Audit observations are assigned to a pocket in each
map (in the forest map, weak or ambiguous assignments abstain). For each pocket the audit estimates the mean
prediction error $\hat\mu_c$ and its between-group gap $\hat\delta_c$; each map's tested pockets receive
simultaneous intervals (Section~\ref{sec:inference}), and an observation inherits its pocket's flag in either map. The
similarity forest $g$ is distinct from the assessed models $f$ (any model or score).}
\label{fig:pipeline}
\Description{A flow diagram: data are split by unit into development, discovery and audit roles; discovery
produces two frozen pocket maps; audit observations are assigned to pockets in each map, each pocket's mean error
and between-group gap are estimated with model predictions, and the report gives flags with simultaneous intervals,
errors ruled out and untested pockets.}
\end{figure}

This section builds the pocket maps of Figure~\ref{fig:pipeline}; Section~\ref{sec:inference} sets what is
estimated in them.

\phead{Setting.} An \emph{assessed model} $f$ maps the features $x$ of an observation to a prediction $f(x)$
of an outcome $Y$: a probability for a binary outcome or a predicted value for a continuous one. The audit requires
features, outcomes, each model's predictions and identifiers of independent sampling units, but not any model's
internals; a between-group contrast also needs a recorded attribute $S$ that defines a \emph{focal} group ($S=1$)
and a \emph{reference} group ($S=0$). We take
$S$ binary for simplicity; a multi-category attribute is audited as pre-declared focal--reference pairs whose
intervals hold jointly (Section~\ref{sec:inference}). A \emph{pocket map} $C(x)$ sends each observation to one of a finite set of pockets or, if its assignment is weak or ambiguous, to ``abstain''.
Neither $S$ nor any model's errors are inputs to the models that form the pockets, although correlated features
can still give pockets different group compositions. Each row is an observation, such as a patient-year, an image or
a transaction; rows come in independent \emph{units} $h$ of possibly correlated observations, such as hospitals.

\phead{Three data roles.} An audit is only as credible as the separation between the data that shape it and
the data that test it \cite{cox1975note}. Whole units are assigned at random, with a fixed seed, to
\emph{development} (40\%), where assessed models are fit if needed; \emph{discovery} (30\%), where pockets are
learned; and \emph{audit} (30\%), where errors are estimated (Figure~\ref{fig:pipeline}). Discovery units are split
once more: 80\% supply the \emph{graph observations} from which pockets are built and 20\% calibrate the rule that
assigns new observations to pockets. Unlike a train/test split, no role needs most of the data:
development fits models, discovery forms pockets and the audit tests them. Audit outcomes are read once, after the models, maps, support rule and inference settings are fixed. We do
not rotate the roles, so every audit statement concerns the audit sample.

\phead{Forest similarity.} A random forest $g$ is grown on the graph observations to predict the outcome. For
observations $i$ and $j$ with leaves $l_t(i)$ and $l_t(j)$ in tree $t$ of $T$, their similarity is the number of
trees in which they share a leaf, $s_{ij} = \sum_{t=1}^{T} \mathbf{1}\{l_t(i) = l_t(j)\}$. Because splits are chosen
to predict the outcome, two observations that differ only in a feature that never predicts it \crx{tend to share}
leaves \cite{lin2006random}; a finite forest can split on noise, so \crx{some trees may separate them}. Categorical
features enter as integer codes, an arbitrary order that remains a choice. We use 200 trees, at least 20 observations per leaf and $\sqrt{p}$ candidate features
per split ($p$ features; 500 trees and $p/2$ in continuous simulations).

\phead{From similarity to pockets.} Storing $s_{ij}$ for every pair needs memory quadratic in the number of
observations, so each graph observation is linked to the (approximately) 30 graph observations with \crx{which} it shares
the most leaves, keeping an unweighted link only when the two are in each other's lists. This mutual nearest-neighbor
graph has at most $15n$ links for $n$ graph observations (each keeps at most 30) and damps observations that are close to everyone
\cite{maier2009optimal,radovanovic2010hubs}. Pockets are the graph's communities, found with a nested,
degree-corrected stochastic block model (SBM) \cite{karrer2011stochastic,peixoto2014hierarchical}, which explains
the links by placing observations in blocks whose members link alike, chooses the number of blocks by minimum
description length (the most compact description; best of three seeded fits), and returns a hierarchy
from fine to coarse, each level merging blocks of the one below. A finer level localizes an error more sharply but leaves fewer observations to test, so the
level is chosen on discovery data only\crx{. We} take the level with the most pockets projected to meet a stricter
version of the audit's support rule (Section~\ref{sec:inference}): at least 150 audit observations per compared group
(75 in the continuous simulations) and, for a binary outcome, 10 events and non-events expected from the
predictions; ties go to the coarser level. Projections scale the pocket counts of
discovery calibration observations to the audit size; audit outcomes never choose the level (supplement). Two comparison partitions of the same graph appear in Sections~\ref{sec:sims}
and~\ref{sec:apps}: Leiden, a standard community-detection algorithm \cite{traag2019leiden}, and a \emph{tree cut}, which cuts the hierarchy at different depths in different branches, splitting a pocket only
when its children yield more pockets projected to meet support. In the continuous simulations, where the dense similarity matrix
is affordable, spectral clustering replaces the SBM.

\phead{Frozen assignment with abstention.} Audit observations never enter the graph. Each pocket $c$ of $n_c$ graph observations is stored as a \emph{profile}, the count $n[c,t,l]$ of its
members in every leaf $l$ of every tree $t$, and a new observation joins the pocket with the highest \emph{affinity}
$a_c(x) = T^{-1}\sum_t n[c,t,l_t(x)]/n_c$, its average similarity to the pocket's members. An observation that
resembles no pocket \emph{abstains} if its top affinity, or its lead over the second, falls below the 1st percentile
on calibration observations. Abstaining observations count overall; abstention is reported by group.

\phead{Linear map.} The \emph{ridge map} $C_R$ covers errors concentrated along one strong linear predictor,
which the forest localizes less sharply (Section~\ref{sec:sims}). On the same graph observations, each feature is
median-imputed and standardized, then multiplied by the absolute value of its coefficient in a ridge regression of
the outcome (logistic for a binary outcome, penalty chosen by cross-validation), so that features count in
proportion to their linear relevance. K-means groups the weighted features into as many pockets as the forest map
has, and each audit observation joins the nearest centre.

\section{What We Estimate and How Sure We Are}
\label{sec:inference}

\phead{Where and for whom.} All quantities condition on the frozen models and pocket maps. For an assessed model
$f$ and a pocket $c$ of either map, the \emph{local calibration error} and the \emph{local calibration gap} are
\begin{equation}
  \mu_c = \E[f(X) - Y \mid C = c], \qquad
  \delta_c = \E[f(X) - Y \mid C=c, S=1] - \E[f(X) - Y \mid C=c, S=0],
\end{equation}
where $\E$ is a population mean, $X$ an observation's features and $C = C(X)$ its pocket. $\mu_c > 0$ means the
model over-predicts on average in pocket $c$ (\emph{where}); $\delta_c > 0$ means it over-predicts more, or
under-predicts less, for the focal group than for the reference group there (\emph{for whom}). For a binary outcome,
$0.05$ is five percentage points; otherwise errors are in outcome units. Both are pocket averages: opposite errors
can still cancel inside a pocket, and $\mu_c = 0$ is not calibration at every predicted value. The pocket ``all''
(every audit observation, abstainers included) gives the usual overall error and group gap, so each audit also
reports the result it refines. When two models $f$ and $f'$ are scored on the same observations, the \emph{paired
gap} $\kappa_c = \delta_c(f) - \delta_c(f')$ asks whether one model's gap is larger. A set of quantities audited
together is a \emph{family}; its \emph{target} $\gamma$ is the allowed chance of any false flag in it (5\% here),
and the \emph{nominal level} $\alpha$ of the intervals below is a setting calibrated to meet that target. For each map, $\mu$, $\delta$ and $\kappa$ for every declared model and group contrast form one
family; each analysis below states its family, and the simulations audit $\delta$ alone or $\mu$ alone.

\phead{Support.} A quantity is tested only if each cell it compares (a pocket, or a pocket and group) has at
least 100 audit observations (50 in the continuous simulations) and, for a binary outcome, at least 10 \emph{expected} events and non-events (sums of the
predicted probabilities and their complements), never audit outcomes. With hospitals as units, each cell must also draw on at least 15
distinct and 10 \emph{effective} hospitals ($(\sum_h n_h)^2/\sum_h n_h^2$, where $n_h$ counts the cell's rows from
hospital $h$: the number of equal-sized hospitals that would be as concentrated), and no hospital may hold more than 20\% of the cell, because cluster-robust standard errors are
unreliable with few or dominant clusters \cite{cameron2008bootstrap,mackinnon2017wild}. Pockets that fail the rule
stay in the map, are reported as \emph{untested}, and are never merged to force a test; an estimate with \crx{zero}
\crx{empirical variance} is reported separately\crx{, except a floored binary $\mu$ (below)}. Because the rule never
\crx{reads audit outcomes, findings cannot steer the eligible set}. The cutoffs were fixed in advance, not tuned (supplement).

\phead{Simultaneous intervals.} Each estimate is a signed sum of cell means. Its standard error, the estimated spread
of the estimate over repeated samples, is built from contributions summed within each unit, which allows observations in the same unit to be correlated
\cite{hansen2019clustered}. To control the chance of \emph{any} false flag in the family, every estimate, divided by its
standard error, is compared with the largest such absolute ratio expected anywhere in the family, a single-step max-$t$ procedure
\cite{westfall1993resampling}. Its reference distribution comes from a wild cluster bootstrap
\cite{cameron2008bootstrap,mackinnon2017wild}: in each of $B$ draws ($B = 4{,}999$ in the applications), every unit
receives one random weight (Mammen's two-point weights \cite{mammen1993bootstrap}) that perturbs all its residuals
together, the residuals are centered at the null hypothesis that every tested quantity is
zero, and both the estimate and its standard error are recomputed. The critical value $q$ is the $1-\alpha$ quantile
of the per-draw maxima of the absolute standardized estimates; the interval for quantity $j$, with estimate $\hat\vartheta_j$, is $\hat\vartheta_j \pm q\,\widehat{\mathrm{se}}_j$, and it is
\emph{flagged} when the interval excludes zero (step-by-step formulas in the supplement). Two refinements address
rare binary outcomes. A cell that happens to draw few events looks both biased and falsely precise, so the standard
errors of binary $\mu$ and $\delta$ are floored at the value the predictions imply under no error ($\kappa$, in which
the outcome cancels, is not floored); the floor \crx{stabilizes the standard error and does not model}
dependence. And because the nominal level need not equal the actual false-flag rate, it is calibrated by simulation: we use
$\alpha = 0.025$ with independent units and $0.0125$ with hospital units, fixed in development simulations before
confirmation and real-data audits, to meet $\gamma = 5\%$; these are settings evaluated for our binary $\delta$ designs, not validated defaults
for arbitrary dependence, and a dependent-unit $\mu$ design exceeded the 5\% target at them (Section~\ref{sec:sims}). Continuous outcomes use $\alpha = 0.05$ without the floor.

\phead{Ruling errors out.} A pocket that is not flagged is not evidence that the model is calibrated there: its
interval may simply be wide. To make the absence of flags informative, we report the \emph{equivalence share}
$E(t)$, the share of all focal audit observations (abstained and untested included in the denominator) whose pocket
was tested and whose simultaneous interval lies within $(-t, t)$ (for $\mu$, over all observations). When the
intervals cover their targets, it rules out large pocket-\emph{average} errors, not errors within a pocket; whether $t$ is small
enough is a judgement for the application.

\phead{Combining the two maps.} The maps are not averaged and no estimate is pooled across them. Each map is
audited as its own family with its own bootstrap, at half the nominal level, $\alpha/2$. An observation receives a
directional flag when either map flags its pocket; when the two maps flag opposite directions, it is recorded as a
conflict instead. If each map's actual chance of any false flag is at most $\gamma/2$, the
Bonferroni inequality bounds the chance across both maps by $\gamma$, whatever the overlap between the maps. Our
bootstrap supplies simulation evidence for this premise, not a finite-sample guarantee, so we evaluate the combined
procedure directly (Section~\ref{sec:sims}). The price of the split is a slightly higher bar in each map; the gain is
that an error along one strong linear feature and an error with curved or interacting structure can both be found
by the same audit.

\section{Simulations}
\label{sec:sims}

\begin{figure}[t]
  \centering
  \includegraphics[width=\linewidth]{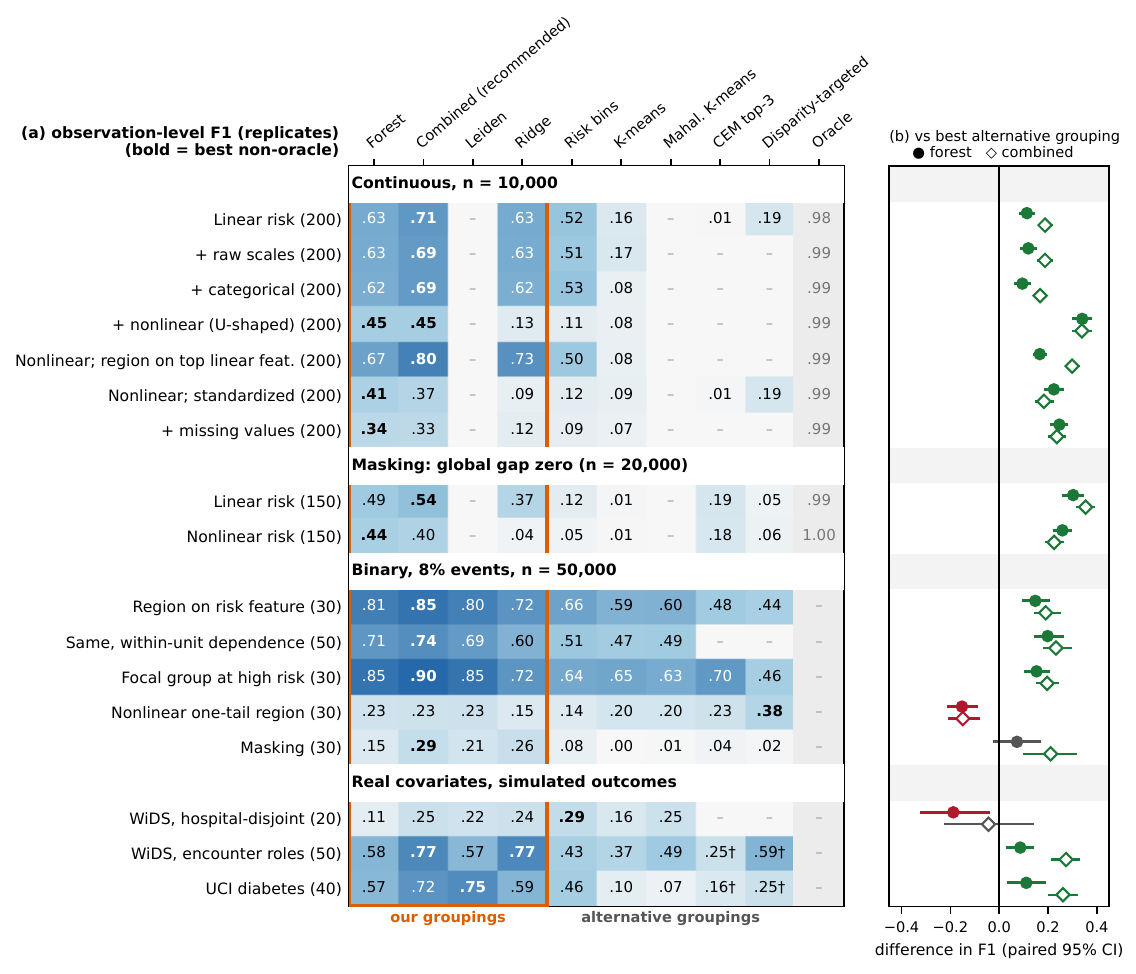}
  \caption{Localization of planted between-group gaps ($\delta$) in 17 simulation scenarios. (a) Observation-level $F_1$ for each grouping, with the number of
  replicates; bold marks the best non-oracle \crx{score in the row and any within 0.005}; outlined columns are our groupings. ``Forest'' is
  the spectral partition of the forest similarity in the continuous rows and the SBM partition of the forest graph
  elsewhere; ``Combined'' is the forest + ridge audit; ``Leiden'' partitions the forest graph with Leiden; ``Ridge''
  is ridge-weighted K-means. Where a scenario ran in two batches, the displayed batch was selected retrospectively by
  a fixed rule, the batch in which the strongest \crx{single grouping other than Forest, Leiden and Oracle} scored highest; $^\dagger$: shown
  from the other batch and left out of panel (b); --: not run. (b) Paired differences with 95\% replicate-bootstrap
  intervals against the strongest available alternative grouping (mostly risk bins; named per row in the supplement):
  dots, forest; diamonds, combined audit. Intervals entirely right of zero favor our audit (green), entirely
  left favor the alternative (red); they are pointwise, and the 17 comparisons are not simultaneous.}
  \label{fig:scoreboard}
  \Description{A heat map of F1 scores, 17 scenario rows by ten method columns, with our four groupings outlined, and
  a dot plot of paired differences against the strongest alternative grouping, with confidence intervals.}
\end{figure}

Simulations provide known prediction errors against which localization can be scored. We ask whether the audit
stays quiet when there is no error (Table~\ref{tab:errors}), how well its flags recover the
observations that carry a planted between-group gap (\emph{for whom}; Figure~\ref{fig:scoreboard}) or a planted local
miscalibration (\emph{where}; Table~\ref{tab:mu}), and when outcome-learned similarity wins.

\phead{Design.} Every replicate of the $\delta$ designs reruns the whole pipeline: new outcomes (and, in the synthetic designs, new
features), a new role split, a new forest and pockets, frozen assignment, and the audit. The assessed model is the
true risk (event probability, or conditional mean) plus a planted error: focal-group observations in a region $R$
receive a prediction shifted by a fixed amount (clipped to valid probabilities for binary outcomes), and everyone
else is predicted exactly. In the synthetic designs the
region covers 20\% of observations and the focal group 20--30\%. Four families of scenarios form the rows of
Figure~\ref{fig:scoreboard}: (i) a continuous outcome, $n = 10{,}000$, with 12 numeric features (plus 3 categorical
in five of the seven specifications) and a shift of 0.4 outcome standard deviations, in a ladder that adds one complication at a time
(raw feature scales, categorical features, a U-shaped risk term with the region on that feature's tail, missing
values), with two controls that place the region on the strongest linear feature or remove the transformations;
(ii) \emph{masking}, $n = 20{,}000$, where the focal group's shift is positive in one half of the region and negative
in the other, so that the population-wide gap is zero; (iii) a rare binary outcome, as in clinical risk models, $n = 50{,}000$ with 8\% events
and a 10-point over-prediction, with variants that add within-unit dependence across 150 units, make the focal group
more common among high-risk observations, place the region on one tail of a U-shaped feature, or mask; and
(iv) real covariates, for realistic feature structure, from the ICU data of Section~\ref{sec:apps} (WiDS; roles
split by stay or by hospital, \emph{hospital-disjoint}) and diabetes data (UCI; supplement), with outcomes simulated from a model fit to the real data and a 10-point error
planted in a region fixed in advance (ventilated patients; two or more prior inpatient stays). These scenarios evaluate
localization of the between-group gap $\delta$; a separate re-audit of saved replicates evaluates $\mu$ (below).

\phead{Comparators and scoring.} Every comparator is learned on the same discovery data, assigns audit observations
with its own frozen rule, and goes through the same support rule and inference, so only the grouping differs.
Methods with a pocket count target the forest map's count in that replicate.
\emph{Our groupings} are the forest pockets, the same forest graph partitioned by Leiden, and the ridge map alone
(\emph{ridge-weighted K-means}). The \emph{alternative groupings} are K-means on standardized features; Mahalanobis
K-means; \emph{risk bins} of the similarity forest's predicted risk; \emph{CEM top-3}, coarsened exact matching (which forms
its own bins) on the three features the forest ranks most important; and a \emph{disparity-targeted} method that learns, on discovery data, where the
assessed model's errors differ between groups and bins that prediction. A truth-aware \emph{oracle} handed the
planted region is shown as a benchmark, not a competitor. Risk bins and CEM top-3 thus borrow the similarity forest,
and the disparity-targeted method uses the group attribute and the model's errors, which our maps never do. These
comparisons isolate the grouping within a common audit; they do not establish superiority over published
slice-discovery systems. Scoring concerns focal-group observations, which are \emph{affected} if they carry the
planted error. A correct-direction flag on an affected observation is a true positive; a flag on an unaffected
observation is a false positive, and a wrong-direction flag on an affected one is both a false positive and a miss;
affected observations in unflagged or untested pockets are misses (FN), and conflicting flags from the two maps count as
errors. We report observation-level
$F_1 = 2\,\mathrm{TP}/(2\,\mathrm{TP}+\mathrm{FP}+\mathrm{FN})$ from counts summed over replicates, with paired, pointwise 95\%
intervals from a bootstrap that resamples the same replicates for every method.

\phead{Error control.} Table~\ref{tab:errors} reports how often the audit flags any pocket when no error exists.
For $\delta$, the two 450-replicate binary designs stayed well below 5\% at their calibrated levels, and the continuous designs, at
nominal 5\%, had intervals that include 5\%. For $\mu$ with independent observations, the combined audit flagged
3.1--5.3\% of null replicates in the 150--500-replicate designs, and each map at most 3.3\% under a weak null whose errors cancel inside every pocket.
With 150 dependent units of unequal size, it flagged 41 of 450 (9.1\%; Wilson interval 6.8--12.1\%) at the level
calibrated for $\delta$, and 9--11\% per map under the weak null; a retrospective estimate, not yet confirmed on fresh
replicates, puts the needed single-map level near 0.0025 (supplement). Plausibly, a pocket's mean error, unlike a
within-pocket gap, does not difference out shared unit-level variation, so $\mu$ under strong within-unit dependence
needs its own level calibration.

\phead{Coverage and checks on real data.} \emph{Simultaneous coverage} is the share of replicates in which every
interval contains its true value. Under a planted gap, among replicates with a tested quantity, forest-map coverage
ranged from 138 of 150 replicates in linear masking (92.0\%; Wilson interval 86.5--95.4\%) to
100\% in the synthetic designs, against a 95\% target, and was 13 of 15 in hospital-disjoint WiDS (86.7\%;
62.1--96.3\%); these checks do not establish nominal coverage in every setting. Under a planted local
miscalibration at the smaller shift of the $\mu$ designs (below), simultaneous coverage of the pocket errors in
the audit sample, among evaluable runs, was 94.0--97.5\% for the combined audit and 93.0--97.5\% for the forest map. Before
the care-management and ICU audits of $\mu$,
\emph{conditional checks}, which simulate outcomes for the real audit observations in their frozen pockets (500
replicates per scenario; supplement), gave
false-flag rates of 0--0.4\% per map in the ICU data (level 1.25\%) and 1.0--4.0\% for the combined care-management
audit, whose family also holds $\delta$ and $\kappa$, with simultaneous coverage of at least 96.0\%.

\begin{table}[t]
\caption{False flags when no error exists: the share of replicates in which any pocket (or the overall comparison)
is flagged, with 95\% Wilson (binomial) intervals (rows of seven specifications: ranges, and the envelope of their intervals).
The $\delta$ rows audit between-group gaps; the $\mu$ rows audit local errors alone, in the localization simulations.
``Forest'' is audited at the single-map level; ``Combined'' audits each map at half of it. Binary levels are
simulation-calibrated for $\delta$ (Section~\ref{sec:inference}); in the $\delta$ 150-unit design, 50 of 450 forest-map
replicates had nothing testable and count as unflagged. No row shows clear inflation above 5\% except $\mu$ with 150 dependent units; the 30-replicate
small check is imprecise.}
\label{tab:errors}
\small
\begin{tabular}{@{}llrll@{}}
\toprule
Null design & Single-map $\alpha$ & Replicates & Forest & Combined \\
\midrule
\multicolumn{5}{@{}l}{\emph{Between-group gap $\delta$}} \\
Continuous, $n = 10{,}000$, 7 specifications & .05 & 700--900 & .050--.059 [.036, .076] & .046--.057 [.033, .074] \\
Continuous, $n = 20{,}000$, linear risk & .05 & 700 & .059 [.043, .078] & .049 [.035, .067] \\
Continuous, $n = 20{,}000$, nonlinear risk & .05 & 700 & .054 [.040, .074] & .047 [.034, .065] \\
Binary, $n = 50{,}000$, independent observations & .025 & 450 & .016 [.008, .032] & .020 [.011, .038] \\
Binary, $n = 50{,}000$, 150 dependent units & .0125 & 450 & .013 [.006, .029] & .013 [.006, .029] \\
Binary, focal group at high risk (small check) & .025 & 30 & .000 [.000, .114] & .033 [.006, .167] \\
\addlinespace
\multicolumn{5}{@{}l}{\emph{Local error $\mu$}} \\
Continuous, $n = 10{,}000$, 7 specifications & .05 & 500 & .036--.060 [.023, .084] & .034--.046 [.021, .068] \\
Continuous, $n = 20{,}000$, linear risk & .05 & 150 & .060 [.032, .110] & .047 [.023, .093] \\
Continuous, $n = 20{,}000$, nonlinear risk & .05 & 150 & .053 [.027, .102] & .053 [.027, .102] \\
Binary, $n = 50{,}000$, independent observations & .025 & 450 & .029 [.017, .049] & .031 [.019, .052] \\
Binary, $n = 50{,}000$, 150 dependent units & .0125 & 450 & .089 [.066, .119] & .091 [.068, .121] \\
\bottomrule
\end{tabular}

\end{table}

\phead{Localizing for whom ($\delta$).} Figure~\ref{fig:scoreboard}a gives each method's $F_1$ in the 17
scenarios and Figure~\ref{fig:scoreboard}b the paired difference between our audit and the strongest available
alternative grouping in the same replicates. We call a difference a win or a loss when its paired interval excludes
zero, and otherwise say the methods are not distinguishable, which is not evidence that they are equivalent. The
combined audit beat the strongest alternative in 15 scenarios, by 0.17--0.35 in $F_1$; it was not distinguishable in
hospital-disjoint WiDS, where support is thin, and lost in the one-tail scenario. The forest pockets alone won 14,
were not distinguishable in 1 (binary masking) and lost 2. Against the best single grouping of any kind, ours
included, the combined audit won 8, was not distinguishable in 6 and lost 3.

\phead{When each grouping wins.} Learned similarity pays off when the structure that matters is not a straight
line in the features: with a U-shaped risk term the forest pockets reached $F_1$ 0.34--0.45 against 0.09--0.13 for
ridge-weighted K-means and risk bins, and under masking they separated the two halves (0.49 and 0.44, against 0.37
and 0.18). With linear risk the two maps scored alike (0.62--0.63). The forest lost when the region is the top of
the most predictive linear feature, which ridge-weighted K-means slices exactly (0.73 vs 0.67; WiDS ventilated
patients, 0.77 vs 0.58); when only one tail of a U-shaped feature is affected, where its pockets mix affected and
unaffected observations and the disparity-targeted method wins (0.38 vs 0.23); and with hospital-disjoint roles in
WiDS, where most affected observations sit in untested pockets ($F_1 \le 0.29$ for every method). The two maps
fail in different places, which motivates running both. Against its two component maps, the combined audit was better in 9 scenarios, not
distinguishable in 6 and worse in 2: each map finds affected observations the other misses, and halving each map's
level costs 3--4 points where the forest alone carries the signal. The combined audit was specified after the
single-map runs;
fresh continuous seeds confirmed the pattern (supplement), and the binary and real-covariate rows reuse saved
replicates retrospectively.

\phead{Localizing where ($\mu$).} We re-audited saved replicates of 15 scenarios with an error planted for
every observation in a region, whatever its group, and audited $\mu$ alone on the saved maps (supplement). Two
scenarios move the region to a feature unrelated to the outcome. The alternatives are risk bins, K-means,
Mahalanobis K-means, CEM and CEM top-3, as available in each batch, and an \emph{error-targeted} grouping refit for
each condition, the $\mu$ analogue of the disparity-targeted method: like slice-finding methods, it bins a boosted
model of one model's residuals learned on discovery data.
At the smaller shift (0.2 outcome standard deviations or 5 points), paired pointwise intervals favored the combined
audit over every available alternative in 12 of 15 scenarios (Table~\ref{tab:mu}), including masking designs with
offsetting local errors and near-zero overall error, where the overall comparison flagged at most 0.7\% of
replicates while the combined audit reached $F_1$ 0.45--0.62. It lost to the error-targeted grouping in the two
irrelevant-feature scenarios (0.17 against 0.07; 0.75 against 0.43) and the binary one-tail region (0.69 against
0.28). At the $\delta$ designs' larger shift, it beat every available alternative in 8, lost to at least one in
6 and was not distinguishable in 1: the better grouping depended on the error.

\begin{table}[t]
\caption{Localizing a planted local miscalibration $\mu$: observation-level $F_1$, scored over all observations
(Figure~\ref{fig:scoreboard} scores focal-group observations), of the combined audit, the error-targeted grouping and
the best other alternative, chosen separately at each shift (named at the smaller one; CEM-3 is CEM top-3); shifts of 0.2 SD or 5 points
(smaller) and 0.4 SD or 10 points (the $\delta$ designs). Bold: the largest unrounded score of the three. Null flags:
replicates with no planted error in which the combined audit flagged any $\mu$; 9.1\% (150 dependent units) exceeds the 5\%
target, and dashes mark irrelevant-feature rows, whose null runs are those of the matching risk-feature design. Binary
one-tail and masking rows reuse maps frozen under the $\delta$ design (30--40 replicates); others use 150--500. Two
intermediate continuous specifications, which behave like linear risk, and all other groupings are in the
supplement.}
\label{tab:mu}
\small
\setlength{\tabcolsep}{4pt}
\begin{tabular}{@{}lrrlrrrr@{}}
\toprule
 & \multicolumn{3}{c}{Smaller shift} & \multicolumn{3}{c}{Larger shift ($\delta$ designs)} & Null \\
\cmidrule(lr){2-4}\cmidrule(lr){5-7}
Scenario & Comb. & Error-t. & Best other & Comb. & Error-t. & Best other & flags, \% \\
\midrule
\multicolumn{8}{@{}l}{\emph{Continuous outcome, $n = 10{,}000$}} \\
Linear risk & \textbf{.73} & .19 & .55 (risk bins) & \textbf{.71} & .56 & .64 & 3.6 \\
\quad + scales, categories, U-shaped risk & \textbf{.56} & .18 & .10 (risk bins) & \textbf{.60} & .57 & .30 & 3.8 \\
Nonlinear; region on top linear feature & \textbf{.80} & .20 & .52 (risk bins) & \textbf{.78} & .57 & .67 & 3.8 \\
Nonlinear; standardized & \textbf{.53} & .18 & .10 (risk bins) & .51 & \textbf{.57} & .30 & 4.6 \\
\quad + missing values & \textbf{.44} & .14 & .09 (risk bins) & \textbf{.50} & .48 & .27 & 4.4 \\
Linear; region on an irrelevant feature & .07 & \textbf{.17} & .17 (K-means) & .23 & \textbf{.57} & .40 & -- \\
\multicolumn{8}{@{}l}{\emph{Masking: overall error zero, $n = 20{,}000$}} \\
Linear risk & \textbf{.59} & .04 & .21 (CEM-3) & \textbf{.66} & .53 & .60 & 4.7 \\
Nonlinear risk & \textbf{.45} & .04 & .22 (CEM-3) & .49 & .53 & \textbf{.60} & 5.3 \\
\multicolumn{8}{@{}l}{\emph{Binary, 8\% events, $n = 50{,}000$}} \\
Region on risk feature & \textbf{.87} & .69 & .71 (K-means) & \textbf{.84} & .78 & .75 & 3.1 \\
Region on an irrelevant feature & .43 & \textbf{.75} & .74 (Mahal.) & .35 & \textbf{.85} & .75 & -- \\
Within-unit dependence & \textbf{.41} & .31 & .32 (K-means) & \textbf{.71} & .66 & .59 & 9.1 \\
Nonlinear one-tail region & .28 & \textbf{.69} & .28 (K-means) & .34 & \textbf{.86} & .38 & 0.0 \\
Masking & \textbf{.62} & .22 & .15 (risk bins) & .68 & \textbf{.73} & .55 & 3.3 \\
\bottomrule
\end{tabular}

\end{table}

\phead{Settings and cost.} The number of candidate features per split changes which structure the forest
emphasizes, in ways the forest's own held-out accuracy does not predict (supplement), so settings are fixed in advance. On discovery graphs of
about 22{,}000 observations the nested SBM took 3--9 minutes and under 2.5~GB; $B = 4{,}999$ draws take seconds.

\section{Applications}
\label{sec:apps}

We demonstrate the audit on two models with documented problems: a commercial care-management score whose cost
basis understated Black patients' health needs \cite{obermeyer2019dissecting}, read as a prediction of illness in
public synthetic data, and the APACHE~IVa \crx{ICU severity score, whose over-prediction of hospital mortality} in recent cohorts has been
reported \cite{sarkar2021apache}, asking in each what a local audit adds to what is known. The first shows
offsetting local errors and one map serving three models; the second, a black-box score with hospital dependence. A briefer third audit, of two chest X-ray classifiers, shows the
method where pockets are learned from image features.

\phead{A care-management algorithm.} \citet{obermeyer2019dissecting} showed that a commercial algorithm used to
enroll patients in care-management programs predicted cost, not health, so that Black patients were sicker than
White patients with the same score, and that training on a health label reduced the gap. It is a widely studied
case of a deployed model that erred between groups, and its authors released person-level synthetic data
generated from the original records: 48{,}784 patient-years (11\% Black). The synthesis does not keep
patients' records together, so we treat rows as independent units; the intervals therefore do not carry the original
patients' sampling uncertainty. The outcome is the number of active chronic conditions in the year of the score
(cohort mean 1.35). Pockets are learned from 148 prior-year features (conditions, costs, laboratory tests,
medications, age bands), never race or sex. We audit three predictions of the outcome, each mapped to its scale by an
isotonic (monotone) regression fit on development rows: the commercial score (the \emph{health-mapped score}), our reading of a
cost score as a prediction of illness, not how it was deployed; and two lasso (penalized linear) regressions on the original study's model features,
one of $\log(\text{cost} + 1)$ (the \emph{cost label}) and one of the number of conditions (the \emph{health label}),
the two labels that study compared. The lasso models differ only in their label, so their paired gap $\kappa$
isolates the choice of label for this model class and calibration. The protocol, committed before
any pocket-level analysis, fixed the roles, the combined audit at $\alpha/2 = 0.025$ per map, the
families ($\mu$, $\delta$ and $\kappa$ for Black--White and female--male) and equivalence tolerances $t$ of 0.25 and 0.5 conditions. Pre-specified conditional checks of false
flags and coverage passed before the audit (Section~\ref{sec:sims}). The forest map
has five pockets, named from the typical features of their discovery members; 2\% of audit rows abstained.

\begin{figure}[t]
  \centering
  \includegraphics[width=\linewidth]{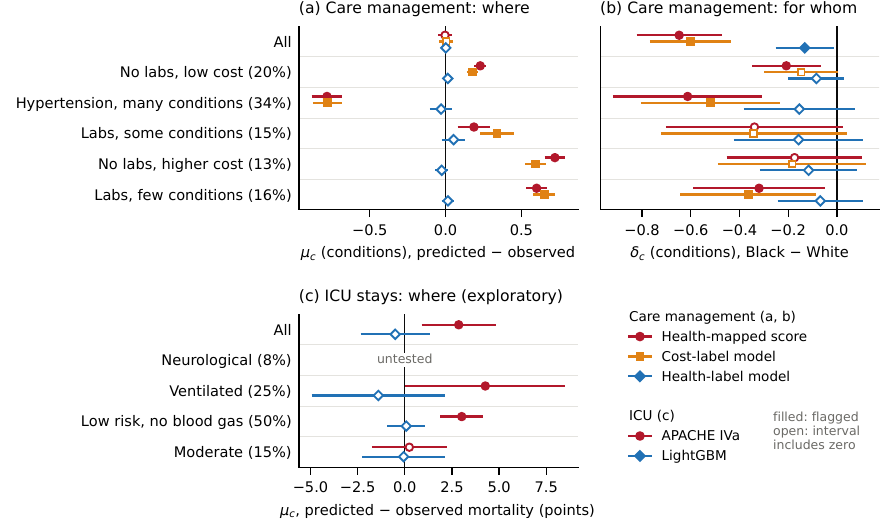}
  \caption{Overall errors conceal opposing local errors; the health-label model has smaller errors and gaps.
  (a) Local calibration error $\mu_c$ of three predictions of the number of chronic conditions, by forest pocket of the
  care-management data. (b) The Black--White gap $\delta_c$ in the same pockets.
  (c) Exploratory local calibration error of APACHE~IVa and LightGBM \crx{predictions of hospital mortality for ICU stays}, by tree-cut pocket.
  Bars are simultaneous intervals (care management: forest map at $\alpha/2 = 0.025$; ICU: $\alpha = 0.0125$);
  filled markers are flagged (open: interval includes zero); percentages give each pocket's share of all audit
  observations (about 2\% abstained and belong to no pocket; ICU: APACHE-valid stays). Pocket names summarize
  the typical features of discovery members, not exact membership rules. The neurological pocket could not be tested because one
  hospital held more than 20\% of its stays.}
  \label{fig:apps}
  \Description{Three forest plots. Panel a: the commercial score and cost-label model under-predict chronic
  conditions by about 0.8 in the hypertension pocket and over-predict by 0.2 to 0.7 in four other pockets, while the
  health-label model is near zero everywhere. Panel b: Black minus White gaps are negative for the cost-based
  predictions, largest in the hypertension pocket, and small for the health-label model. Panel c: APACHE IVa
  over-predicts mortality overall, in ventilated and in low-risk stays, but not in a moderate pocket; LightGBM is near
  zero.}
\end{figure}

\emph{Where.} The health-mapped score showed no detectable overall error in
the audit sample (1.35 conditions predicted, 1.36 observed; $\mu = -0.003$ $[-0.049, 0.042]$), yet it erred in every forest
pocket (Figure~\ref{fig:apps}a). It under-predicted by 0.78 conditions $[0.68, 0.88]$ in the pocket of patients with
hypertension and many prior conditions, a third of patient-years, and over-predicted by 0.19--0.72 in the four
healthier pockets, most for patients without routine laboratory tests but with higher prior costs (0.92 predicted,
0.20 observed). The score compresses illness toward the mean, and the overall check averages the two errors away. The
ridge map, which slices along the number of prior conditions, found the extreme: in the pocket of the 3\% of
patient-years with the most prior conditions and renal disease, the score predicted 4.3 conditions against 7.7
observed ($\mu = -3.4$ $[-4.0, -2.7]$). The cost-label model erred in the same places. The health-label model's intervals bounded every
forest-pocket error within $\pm 0.25$ conditions; no pocket of either map was flagged.

\emph{For whom.} Overall, the health-mapped score under-predicted Black patients' conditions by 0.65 more than
White patients' ($\delta = -0.65$ $[-0.82, -0.47]$), in the direction of the original finding. The pockets show two patterns in
these synthetic data, matching the two ways a gap can arise (Section~\ref{sec:intro}). Black patients were over-represented where the score under-predicted on average in both groups: 49\% of
Black patient-years, against 33\% of White, were in the hypertension pocket, and in the combined audit 50\% of Black
against 33\% of White patient-years were in a pocket flagged for under-prediction (the two maps flagged opposite
directions for 2.2\% of rows, reported separately). And a Black--White gap remained within pockets, flagged in three
of five forest pockets and largest in the hypertension pocket ($-0.61$ $[-0.92, -0.31]$; Figure~\ref{fig:apps}b).
For these fitted models, training on the health label substantially reduced the gap: the paired gap between the cost- and health-label models was
$-0.47$ $[-0.58, -0.35]$ overall and was flagged in the hypertension pocket ($-0.36$) and one other. The health-label
model had no flagged pocket gap in either map; in the forest map, 98\% of Black patient-years were in pockets whose
interval excluded gaps beyond $\pm 0.5$ conditions ($E(0.5) = 98\%$), though a small overall gap remained ($-0.13$ $[-0.25, -0.01]$).
The secondary female--male contrast showed what a check of one attribute misses: the cost-label model
over-predicted women's conditions by 0.73 more than men's $[0.64, 0.82]$, flagged in every forest pocket, a gap the
health-mapped score showed only overall ($+0.11$ $[0.02, 0.21]$) and the health-label model did not show. The data are synthetic, so these findings
describe the released data, not the original population.

\phead{A deployed ICU severity score.} The WiDS Datathon 2020 data \cite{lee2020wids}, assembled by the GOSSIS
initiative \cite{raffa2022gossis} and distributed through PhysioNet \cite{goldberger2000physionet}, contain 91{,}713
ICU stays from 147 hospitals, 8.6\% of which ended in death before hospital discharge, with about 90 variables
recorded in the first 24 hours. \crx{Most stays carry} the APACHE~IVa predicted probability of death
\cite{zimmerman2006apache}, which we use as a black box. We also audit a LightGBM (boosted-tree) model \cite{ke2017lightgbm} fit on
development hospitals and recalibrated with Platt (logistic) scaling \cite{platt1999probabilistic}. Hospitals are the units, so
the 44 audit hospitals (23{,}216 stays) neither trained a model nor shaped a pocket; 40 of them contributed the
20{,}923 stays with a valid APACHE probability, used for the comparisons of $\mu$ below. A protocol committed before any audit outcome was read named the SBM forest pockets as the primary map
and a tree cut of the same hierarchy, which yields four clinically nameable pockets, as a pre-declared comparison. It
specified the between-group gaps ($\delta$ and $\kappa$); the local errors $\mu$ below use the tree cut, the finer
map, and were added after that audit sample had been read, so they are exploratory and form a separate family. Conditional checks on the same stays and pockets passed before $\mu$ was computed
(Section~\ref{sec:sims}).

\emph{Where.} Overall, APACHE~IVa predicted 11.0\% mortality against 8.1\% observed, an over-prediction of 2.9
percentage points (simultaneous interval $[0.9, 4.8]$); LightGBM's error was $-0.5$ $[-2.3, 1.3]$. The tree-cut
pockets locate APACHE's error (Figure~\ref{fig:apps}c). It over-predicted in the ventilation pocket (79\% ventilated;
21.2\% predicted against 16.9\% observed; $+4.3$ $[0.03, 8.5]$) and in the low-risk pocket of stays mostly without a
blood gas, half of the APACHE-valid audit stays, where its predictions were more than twice the observed rate (5.4\%
against 2.4\%; $+3.0$ $[1.9, 4.2]$). In a moderate-risk pocket no error was detected (5.7\% against 5.4\%; $+0.2$
$[-1.7, 2.2]$). Over-prediction was detected in two of the three testable pockets at the reported exploratory level. The neurological pocket could not be tested, because one hospital held more than 20\% of its stays.
LightGBM, which had never seen these hospitals, showed no detectable error in the three testable pockets, and in the
low-risk pocket its interval excluded errors beyond $\pm 1.1$ points. Our simulations suggest that $\mu$ with dependent units can need a stricter level
(Section~\ref{sec:sims}); at a single-map level of 0.0025, the overall and low-risk flags would remain (family-adjusted
$p = 0.002$ and $0.0002$) and the ventilation flag would not ($p = 0.012$).

\emph{For whom.} The gap analyses of that protocol use each model's eligible audit stays. With an 11\% minority
group and a rare outcome, the discovery data supported only two SBM pockets for the Black--White comparison, and no
gap was flagged for either model in either map. LightGBM's overall gap was $+1.4$ points $[-1.1, 3.8]$, and in the
SBM map 97.6\% of Black audit stays were in pockets whose interval excluded a pocket-average LightGBM gap larger than
5 points. No female--male gap was flagged, and the Hispanic--White contrast could not be tested, because Hispanic
patients were concentrated in about four effective hospitals. Results for the pre-declared comparison maps are in
the supplement.

\phead{A chest X-ray classifier.} Under its own frozen protocol (supplement)\crx{, with each map at $\alpha/2 = 0.025$ after conditional checks}, we audited two public chest X-ray
classifiers \cite{cohen2022xrv}, recalibrated on development images, against the report-derived label for effusion (fluid around the lung) in NIH
ChestX-ray14 \cite{wang2017chestxray}, one image per patient. As in the care-management data, neither showed a
detectable overall error, yet both under-predicted by 23--28 points in a pocket of 3\% of images and over-predicted by
5--8 points in a pocket whose discovery images were 95\% anteroposterior (AP) views. No female--male gap was flagged, but local sex-gap tests were
unavailable for 53\% of women's and 41\% of men's images in the forest map (supplement).

\section{Discussion and Limitations}
\label{sec:discussion}

Outcome-learned pockets give reusable maps, the same for every model audited, with unit-aware uncertainty
and an explicit report of where the data are too thin. Each part of the design guards against a different failure:
learning pockets from the outcome, not from any model's errors, keeps them comparable across models and lets
groups be compared within shared pockets; separating discovery from audit keeps flags from being tuned to the audit
outcomes; and simultaneous intervals with a support rule show how large an undetected error could be and which pockets
could not be tested. The two questions answer different things: $\mu$
locates pocket-average errors, and together with group shares across pockets, $\delta$ distinguishes a group's
concentration in poorly served pockets from gaps within them, as in the care-management data. The forest map helped most with nonlinear or masked
structure, the ridge map with one strong linear feature. Conclusions concern tested pocket averages, given the fitted
models and frozen maps.

\phead{Acting on an audit.} \crx{A flagged pocket is where follow-up starts.} Because the
assignment rule is frozen and needs only features, a new observation in a flagged pocket can be routed for review at
prediction time (the audit gives no individual prediction intervals). The pocket can also guide repair: recalibrating
within it, reweighting it or training robustly over pockets \cite{sagawa2020distributionally}, a separate model, or, as in the care-management example, a label that better
reflects what the decision should track. Each repair must be evaluated on data not used to find the pocket. In our
examples, the hypertension pocket is a priority population for evaluating a health-target model; APACHE~IVa, flagged in two of three testable pockets, shows why a recalibration tuned to the overall error needs pocket-level checks on new data.

\phead{What limits a local audit.} The main limit is support. A rare outcome, a small group and unit-level dependence leave few
testable pockets: in the chest X-ray audit each map tested five local sex gaps per model,
against 11--13 pockets for $\mu$, which pools a pocket's observations; the ICU audit adds hospital-level dependence;
and what was held out (hospitals or
encounters) changed localization far more than the community-detection algorithm did (supplement). With a
count outcome and independent rows, every care-management pocket was testable for $\mu$ and $\delta$.

\phead{Limitations.} The audit compares pocket averages: it is blind to errors that cancel inside a pocket and,
for $\delta$, to a model that is equally miscalibrated for both groups. Its intervals cover the tested pockets of
the frozen maps, conditional on those maps, and say nothing about all possible subgroups. Pockets built from outcome
similarity can miss errors unrelated to the outcome, and they dilute an error confined to observations the forest
does not separate well, such as one tail of a U-shaped feature; in both cases a grouping learned from the model's own
residuals did better in our simulations. With few, unequal units, $\mu$ needed a stricter level than $\delta$. An audit against the recorded outcome cannot detect
bias in the outcome itself \cite{obermeyer2019dissecting}; the care-management example works only because a
better-aligned outcome was available. The conditional checks do not validate population-sampling coverage for the paired gap
$\kappa$. Our binary levels were calibrated by simulation at the applications' scale;
they are evidence for the evaluated designs \crx{and carry no} finite-sample guarantee, and the combined audit was specified
after the single-map confirmation runs. The care-management data are synthetic and drop the original patient
clustering, and the ICU analysis of $\mu$ was added after that audit sample had been read. Recorded race is a social
category: a gap describes how a model's errors fall on recorded groups, not why. Clinical review of the application feature lists and tolerances would
strengthen the applications.

\phead{Future directions.} Swapping the discovery and audit roles would \crx{also audit the discovery sample}; this paper
uses a single split. Pockets from other sources, such as language-model
descriptions of failure regions \cite{eyuboglu2022domino}, could be audited the same way. A shared bootstrap across maps, $\mu$ levels for dependent units and repeated audits need their own error control.

\phead{Code and supplement.} Code, simulation scripts, data-access instructions and the supplement are at
\url{https://github.com/modenesi/pocket-audit}. The supplement holds inference details, further simulations and
validation, the diabetes and chest X-ray applications and data processing.

\phead{Use of generative AI\@.} We used generative AI tools (Claude, Anthropic; Codex, OpenAI) to help \crx{with editing, coding} and analysis; the authors reviewed and take full responsibility for all content.

\begin{acks}
This work was supported by Rocket Companies and the Michigan Institute for Data and AI in Society (MIDAS) at the
University of Michigan. The views expressed are the authors' own and do not represent those of any company or
institution, including the authors' employers and the funders.
\end{acks}

\bibliographystyle{ACM-Reference-Format}
\bibliography{refs}

\clearpage
\setcounter{section}{0}\renewcommand{\thesection}{S\arabic{section}}\renewcommand{\theHsection}{S\arabic{section}}
\setcounter{table}{0}\renewcommand{\thetable}{S\arabic{table}}\renewcommand{\theHtable}{S\arabic{table}}
\setcounter{figure}{0}\renewcommand{\thefigure}{S\arabic{figure}}\renewcommand{\theHfigure}{S\arabic{figure}}
\setcounter{equation}{0}\renewcommand{\theequation}{S\arabic{equation}}\renewcommand{\theHequation}{S\arabic{equation}}
\section*{Supplementary Material}
\pdfbookmark[1]{Supplementary Material}{supplementary}
This supplement provides inference details
(Section~\ref{app:inference}), implementation settings (Section~\ref{app:settings}), two illustrations
(Section~\ref{app:illus}), additional simulation results (Section~\ref{app:extra}), the validation of the local
calibration error $\mu$ (Section~\ref{app:muval}), the diabetes readmission application (Section~\ref{app:uci}),
data processing for every dataset (Section~\ref{app:data}), the $\mu$-localization simulations
(Section~\ref{app:musims}) and the chest X-ray application (Section~\ref{app:cxr}).

\section{Inference Details}
\label{app:inference}

Each coordinate $j$ (a local error $\mu$, a pocket gap, a paired gap or a loss contrast) is a signed sum of cell means,
$\hat\vartheta_j = \sum_k s_{jk}\,\bar z_k$, where $z_i$ is the row's residual $f(x_i) - y_i$ (or a
prediction difference or a loss), $\bar z_k$ its mean over the $n_k$ rows of cell $k$, and $s_{jk} = \pm 1$.
With per-unit contributions $U_{hj} = \sum_k s_{jk} \sum_{i \in h \cap k} (z_i - \bar z_k)/n_k$ over the $H$
audit units, and $\bar U_j$ their mean, the cluster-robust standard error is
$\widehat{\mathrm{se}}_j^2 = \frac{H}{H-1}\sum_h (U_{hj} - \bar U_j)^2$. For binary-outcome calibration gaps $\delta$ (not paired gaps $\kappa$, in which the outcome cancels on
common rows, and not loss contrasts) we use
$\widetilde{\mathrm{se}}_j = \max(\widehat{\mathrm{se}}_j, \mathrm{se}^0_j)$, where
$(\mathrm{se}^0_j)^2 = \sum_k n_k^{-2} \sum_{i \in k} \pi^0_i(1-\pi^0_i)$ and
$\pi^0_i = \mathrm{clip}(f(x_i) - \bar r_c, 10^{-6}, 1 - 10^{-6})$ recalibrates the predictions to the
pocket's pooled residual mean $\bar r_c$, the value both groups share under no gap. A local error $\mu_c$ uses one
pooled pocket cell with sign $+1$; its null centre is zero, and for a binary outcome its floor is
$\mathrm{se}^0_c = \sqrt{\sum_{i\in c}f(x_i)\{1-f(x_i)\}}/n_c$, applied unchanged in every bootstrap draw. Continuous
outcomes use no floor.

In bootstrap draw $b$, each unit receives a Mammen two-point weight $\xi_h$ (mean 0, variance 1, third
moment 1) \cite{mammen1993bootstrap}. Every row in cell $k$ is replaced by
$z_i^* = \tilde m_k + \xi_{h(i)} (z_i - \tilde m_k)$, where the centers $\tilde m_k$ impose each
coordinate's null (for a gap, both cells are centered at their pooled mean; for excess loss, at the
cohort mean). The estimate and its standard error are recomputed, with the same floor where it applies, and the draw
records $T_b^* = \max_j |\hat\vartheta_j^*| / \widetilde{\mathrm{se}}_j^*$. All coordinates of a family share
the weights, so a draw represents shared unit-level variation across overlapping pockets, groups and models; joint
coverage additionally requires that the bootstrap consistently estimate the joint distribution of the centred,
studentized statistics. Shared unit weights represent the dependence, but coordinate-specific null centring can change
the per-unit contributions and their correlations. Asymptotic results for the wild cluster bootstrap of studentized
linear statistics \cite{djogbenou2019asymptotic,canay2021wild} motivate the construction; they do not by themselves
establish coverage for this joint family. Our simulations provide evidence for the evaluated designs, including
partial-null cases in which some coordinates are null and others are not (main text, Section 5, and
Section~\ref{app:muval}); conditional outcome resampling does not validate population-sampling coverage for $\kappa$.
The critical
value $q$ is the $\lceil (B+1)(1-\alpha) \rceil$-th order statistic of $T_1^*, \dots, T_B^*$; intervals
are $\hat\vartheta_j \pm q\,\widetilde{\mathrm{se}}_j$ and adjusted $p$-values are
$(1 + \#\{b : T_b^* \geq |\hat\vartheta_j|/\widetilde{\mathrm{se}}_j\})/(B+1)$. Only per-unit cell sums are
needed, so the cost is $O(B \sum_j H_j)$, where $H_j$ is the number of units contributing to coordinate $j$.

Why these choices: an earlier Gaussian multiplier bootstrap with fixed standard errors
\cite{chernozhukov2013gaussian} flagged 9--12\% of null replicates at nominal 5\% with about 45 audit
hospitals and a rare outcome; recomputing the standard error in each draw, imposing the null, the
expected-events rule, the floor and the calibrated levels brought this within target.

\phead{Where the thresholds come from.} The cutoffs protect against sparse or concentrated samples,
not optimized or justified by application-specific utility. The expected-events rule screens out cells where the models predict only a handful of events or
non-events; the hospital rules keep a cell from resting on a few hospitals, or on one large one,
where cluster-robust standard errors are unreliable \cite{cameron2008bootstrap,mackinnon2017wild}. The
\chg{observation} counts and the projected discovery counts were carried over from earlier versions of the pipeline.
Development simulations then showed too many false flags with an earlier inference procedure; the revision
added the expected-events rule, the standard-error floor below and more conservative bootstrap levels
(this section). All settings were frozen before the fresh-seed simulations and the
real-data audits. Those simulations evaluate the procedure as a whole, with these settings; they do not show
that each cutoff is individually necessary or optimal, and lower cutoffs would buy finer pockets at the cost
of noisier tests.

\section{Implementation Settings}
\label{app:settings}

Table~\ref{tab:settings} lists the settings of the evaluated implementations and the reason for each.

\begin{table}[h]
\caption{Settings of the evaluated implementations, and why ($p$: number of encoded features; $K$:
number of pockets\chg{; a \emph{cell} is a pocket, or a pocket--group combination; effective hospitals are defined in Section~4 of the paper; $B$, bootstrap draws}). All single-map settings were fixed before the
confirmation simulations and the real-data audits. The combined audit was specified afterwards, validated on
fresh continuous seeds and evaluated retrospectively on saved runs (main text, Section 5).}
\label{tab:settings}
\small
\begin{tabularx}{\linewidth}{@{}>{\raggedright\arraybackslash}p{2.5cm}>{\raggedright\arraybackslash}p{6.6cm}>{\raggedright\arraybackslash}X@{}}
\toprule
Component & Setting & Reason \\
\midrule
Roles & 40/30/30\% of units; 80/20 split of discovery & keep model fitting, pocket learning and testing apart \\
Similarity forest & 200 trees (binary, real data) or 500 (continuous); $\geq$20 \chg{observations} per leaf; $\sqrt{p}$ (binary, real data) or $p/2$ (continuous) candidate features per split & outcome-relevant similarity; leaf size keeps leaves stable \\
Graph & mutual top-30 neighbors; bounded candidate search & sparse memory, damps hubs \\
Pockets (binary, real data) & nested SBM, best of 3 fits; level with the most supported pockets: $\geq$150 projected audit \chg{observations} per group and $\geq$10 projected expected events and non-events; $\geq$80\% of graph \chg{observations} profiled & menu of resolutions; testable pockets \\
Pockets (continuous sims) & spectral embedding + K-means, $K\in[2,16]$, same rule with $\geq$75 projected \chg{observations} per group & \chgr{dense discovery matrix affordable at these sample sizes} \\
Abstention & top affinity or margin below the 1st percentile & do not force \chg{observations} into an unlike pocket \\
Ridge map $C_R$ & ridge-weighted K-means, same $K$ as the forest map; nearest centre & covers effects concentrated in one strong linear feature \\
Support (audit) & $\geq$100 \chg{observations} (50 in the continuous simulations) per compared cell; binary outcomes: $\geq$10 expected events and non-events; hospitals: $\geq$15 distinct, $\geq$10 effective, none $>$20\%; UCI: $\geq$100 distinct patients & avoid tests that a few \chg{observations} or one hospital could drive \\
Inference & wild cluster bootstrap max-$t$, $B=4{,}999$ (real data, continuous confirmation) or 999--1{,}999 (other
simulation batches); level $\alpha$ = .025 (\chg{independent observations}) or .0125 (hospitals) for binary outcomes, .05 for continuous; chest X-rays .025 per map in the combined audit & dependence-robust simultaneous intervals; levels calibrated by simulation (main text, Section 4) \\
Combined audit & each map its own family at $\alpha/2$; \chg{observation} flagged if either map flags & one \chg{false-flag} budget across both maps (Bonferroni) \\
\bottomrule
\end{tabularx}
\end{table}

\section{Illustrations}
\label{app:illus}

\begin{figure}[h]
  \centering
  \includegraphics[width=\linewidth]{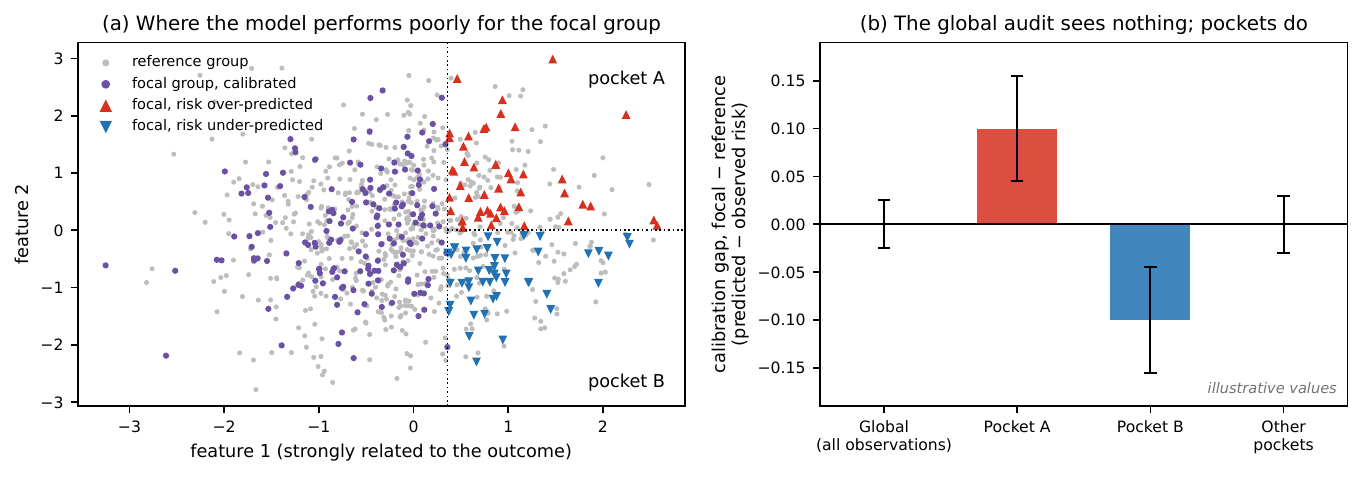}
  \caption{Masking (schematic\chgr{, hypothetical}). (a) The model over-predicts risk for focal-group \chg{observations} in one
  region (pocket A) and under-predicts it in another (pocket B). (b) The two errors cancel in the global
  comparison of focal and reference \chg{observations}; a pocket-level audit sees both. Estimates and error bars are
  illustrative; simulation results for this scenario are in the main text, Section 5.}
  \label{fig:masking}
  \Description{Left: a scatter plot of observations in two features, with focal-group observations over-predicted in
  the upper right corner and under-predicted in the lower right corner. Right: bars of the focal-minus-reference
  calibration gap, zero globally, positive in pocket A and negative in pocket B.}
\end{figure}

\phead{Offsetting errors.} Figure~\ref{fig:masking} shows an extreme, hypothetical case of offsetting errors (called masking in the simulations): errors of opposite sign in two regions cancel in the overall comparison, while a pocket-level audit sees both.

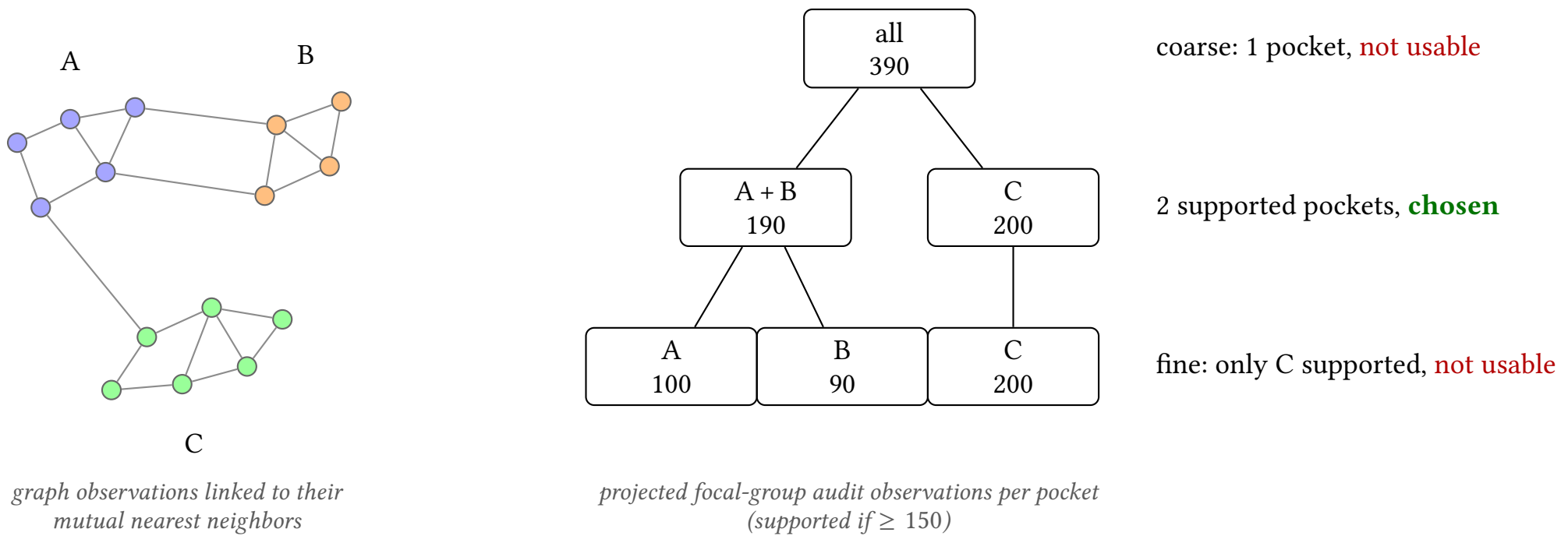
\begin{figure}[h]
\centering
\begin{tikzpicture}[font=\footnotesize,
  pt/.style={circle, inner sep=1.6pt, draw=black!60, fill=#1},
  lv/.style={draw, rounded corners=2pt, align=center, inner sep=3pt, minimum width=1.45cm},
  ok/.style={text=green!45!black}, no/.style={text=red!70!black}]
  \begin{scope}[shift={(0,0)}]
    \foreach \i/\x/\y in {1/0/0, 2/.55/.3, 3/.25/.75, 4/.8/.85, 5/-.2/.55}
      \node[pt=blue!35] (a\i) at (\x,\y) {};
    \foreach \i/\x/\y in {1/1.9/.1, 2/2.45/.35, 3/2.0/.7, 4/2.55/.9}
      \node[pt=orange!50] (b\i) at (\x,\y) {};
    \foreach \i/\x/\y in {1/.9/-1.1, 2/1.45/-.85, 3/1.2/-1.5, 4/1.75/-1.35, 5/.6/-1.55, 6/2.05/-.95}
      \node[pt=green!40] (c\i) at (\x,\y) {};
    \foreach \u/\v in {a1/a2,a2/a3,a3/a4,a2/a4,a1/a5,a5/a3,b1/b2,b2/b3,b3/b4,b1/b3,b2/b4,
                       c1/c2,c2/c3,c3/c4,c4/c6,c2/c6,c1/c5,c5/c3,c2/c4, a4/b3, a2/b1, a1/c1}
      \draw[black!45] (\u) -- (\v);
    \node at (.25,1.25) {A};  \node at (2.25,1.3) {B};  \node at (1.3,-2.0) {C};
    \node[align=center, font=\scriptsize\itshape, text=black!65] at (1.15,-2.55) {graph \chg{observations} linked to their\\mutual nearest neighbors};
  \end{scope}
  \begin{scope}[shift={(5.6,0)}]
    \node[lv] (all) at (1.6,1.35) {all\\390};
    \node[lv] (ab) at (.55,0) {A\,+\,B\\190};
    \node[lv] (c2) at (2.65,0) {C\\200};
    \node[lv] (a) at (-.25,-1.35) {A\\100};
    \node[lv] (b) at (1.2,-1.35) {B\\90};
    \node[lv] (c3) at (2.65,-1.35) {C\\200};
    \draw (all) -- (ab); \draw (all) -- (c2); \draw (ab) -- (a); \draw (ab) -- (b); \draw (c2) -- (c3);
    \node[anchor=west, align=left] at (3.75,1.35) {coarse: 1 pocket, \textcolor{red!70!black}{not usable}};
    \node[anchor=west, align=left] at (3.75,0) {2 supported pockets, \textbf{\textcolor{green!45!black}{chosen}}};
    \node[anchor=west, align=left] at (3.75,-1.35) {fine: only C supported, \textcolor{red!70!black}{not usable}};
    \node[align=center, font=\scriptsize\itshape, text=black!65] at (1.25,-2.55) {projected focal-group audit \chg{observations} per pocket\\(supported if $\geq 150$)};
  \end{scope}
\end{tikzpicture}
\caption{Choosing the audit level from the SBM hierarchy (illustrative counts). Left: graph \chg{observations}
linked to their mutual nearest neighbors in forest similarity fall into three communities. Right: the
hierarchy, with the number of focal-group audit \chg{observations} each pocket is projected to contain (the reference
group and expected events are assumed sufficient). At the finest level only C is supported, so the level is
not usable. Merging A and B gives two supported pockets, so this level is chosen. The coarsest level has one
pocket. No audit outcome is used.}
\label{fig:levels}
\Description{Left: a small network of observations in three clusters labelled A, B and C. Right: a three-level
tree with pockets A (100), B (90) and C (200) at the bottom, A plus B (190) and C (200) in the middle, and all
(390) at the top; the middle level is marked as chosen.}
\end{figure}

\phead{Choosing the audit level.} Figure~\ref{fig:levels} illustrates how the level of the SBM hierarchy is chosen for testability from discovery data only.

\section{Additional Simulation Results}
\label{app:extra}

\begin{figure}[h]
  \centering
  \includegraphics[width=\linewidth]{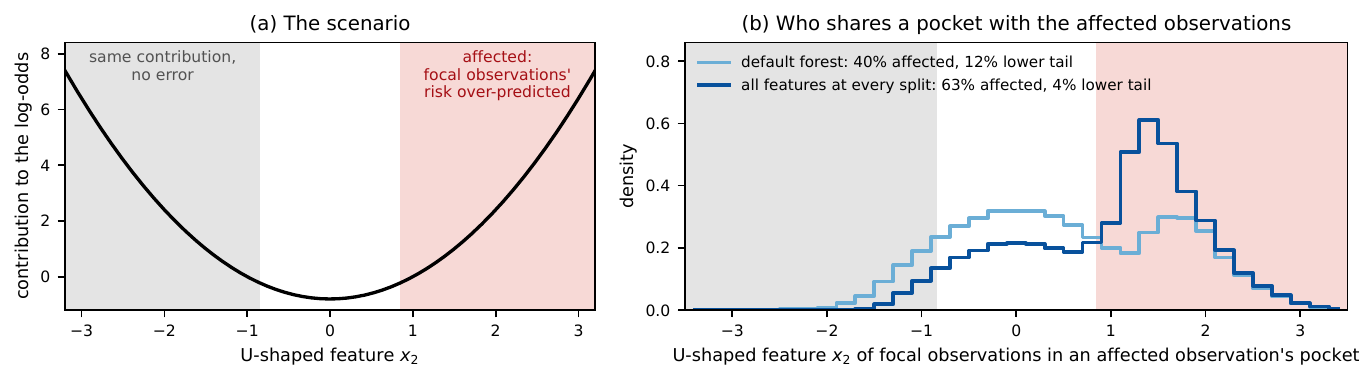}
  \caption{Why the forest only partly finds an affected tail (binary one-tail scenario, 15 replicates paired
  across forest settings). (a) The U-shaped feature raises risk equally at both extremes, but the model
  over-predicts risk only for focal observations in the upper tail. (b) The U-shaped feature among the focal
  observations that share a pocket with an affected observation (pockets weighted by their number of affected
  observations). With the default forest most of them have ordinary values of the feature; with every feature
  considered at every split the pockets concentrate on the affected tail.}
  \label{fig:utail}
  \Description{Left: a U-shaped curve of risk against a feature, with both tails shaded and only the upper
  tail marked as affected. Right: two density curves of the feature among pocket-mates of affected observations;
  the default forest's curve is centred on ordinary values, the all-features curve peaks in the upper tail.}
\end{figure}

\phead{Why the forest only partly finds an affected tail.} Suppose risk rises at both extremes of a feature and the model performs poorly only at the upper extreme (Figure~\ref{fig:utail}a). Risk bins mix the two equally risky extremes, and ridge weighting gives a symmetric U-shaped effect almost no weight, so both localize the affected tail poorly ($F_1$ 0.14--0.15). The forest can, but with default settings only partly: in the pocket of a typical affected observation, only 40\% of the focal observations are affected; 12\% come from the unaffected lower tail, and most of the rest have ordinary values of the feature (Figure~\ref{fig:utail}b). The pocket's gap is diluted accordingly. Considering every feature at every split raises the affected share to 63\% (Table~\ref{tab:mtry}).

\phead{Combined forest + ridge audit.} Table~\ref{tab:union} reports the combined audit of
the main text (Section 5) against each map alone. The continuous rows are a fresh-seed validation, run after
the rule was fixed. The binary and real-covariate rows are a retrospective evaluation: the saved replicates
were re-audited from their stored inputs with each replicate's own bootstrap seed\chgr{, reproducing every stored
single-map count exactly}.

\phead{Forest settings.} Table~\ref{tab:mtry} gives $F_1$ by the number of candidate features per
split, as a share of the $p$ features. Rows pair replicates across settings (development seeds). ``Default''
is $p/2$ in the continuous ladder and $\sqrt{p}$ elsewhere; the 1/3 setting was run only on the real WiDS
covariates.

\phead{Strongest alternative in Figure~2.} The paired differences of Figure~2b are taken against risk bins,
except CEM top-3 in both masking rows of the continuous outcome and the focal-group-at-high-risk row, the
disparity-targeted method in the standardized nonlinear and binary one-tail rows, and Mahalanobis K-means for WiDS
encounter roles.

\phead{Null check at $n = 20{,}000$.} \chgr{Two earlier 150-replicate batches had shown 7--8\% false flags
for the forest pockets; a fresh check with 400 null replicates per scenario gave 5.0\% [3.3, 7.6] in both (ridge
4.3--5.0\%, combined audit 4.5\% and 4.0\%). Table 1 of the main text pools all three batches.}

\begin{table}[!hb]
\caption{The combined audit (forest and ridge-weighted K-means families, each at half the level) against
each map alone at the full level. Observation-level $F_1$; paired 95\% replicate-bootstrap intervals for the
difference from the better of the two maps. Figure~2 of the main text \chg{compares the combined audit with the strongest available alternative grouping} instead, and reports
different batches for the continuous rows and two binary rows.}
\label{tab:union}
\scriptsize
\begin{tabular}{@{}lrcccl@{}}
\toprule
Scenario & Reps & Forest & Ridge & Combined & Combined $-$ better map [95\% CI] \\
\midrule
\multicolumn{6}{@{}l}{\emph{Continuous outcome: fresh-seed validation ($n = 10{,}000$; masking $n = 20{,}000$)}} \\
\quad Linear risk & 200 & .65 & .66 & \textbf{.73} & .07 [.06, .08] \\
\quad + raw feature scales & 200 & .65 & .66 & \textbf{.74} & .07 [.05, .08] \\
\quad + categorical features & 200 & .65 & .67 & \textbf{.74} & .07 [.05, .08] \\
\quad + nonlinear risk & 200 & .43 & .14 & \textbf{.43} & .01 [$-$.01, .02] \\
\quad Nonlinear, region on top linear feature & 200 & .70 & .73 & \textbf{.79} & .06 [.04, .08] \\
\quad Nonlinear, standardized features & 200 & \textbf{.43} & .09 & .39 & $-$.03 [$-$.05, $-$.02] \\
\quad + missing values & 200 & \textbf{.33} & .12 & .32 & $-$.01 [$-$.03, .01] \\
\quad Masking, linear risk & 150 & .45 & .36 & \textbf{.53} & .07 [.05, .10] \\
\quad Masking, nonlinear risk & 150 & \textbf{.44} & .05 & .40 & $-$.04 [$-$.05, $-$.03] \\
\addlinespace\multicolumn{6}{@{}l}{\emph{Binary outcome, $n = 50{,}000$: retrospective, saved replicates (forest = SBM)}} \\
\quad Region on the risk feature & 40 & \textbf{.82} & .70 & \textbf{.82} & .00 [$-$.04, .03] \\
\quad Same, within-unit dependence (150 units) & 50 & .71 & .60 & \textbf{.74} & .03 [.01, .05] \\
\quad Focal group concentrated at high risk & 30 & .85 & .72 & \textbf{.90} & .04 [.01, .07] \\
\quad Nonlinear one-tail region & 40 & \textbf{.24} & .16 & \textbf{.24} & .00 [$-$.02, .01] \\
\quad Masking & 30 & .15 & .26 & \textbf{.29} & .02 [$-$.04, .09] \\
\addlinespace\multicolumn{6}{@{}l}{\emph{Real covariates, simulated outcomes: retrospective, saved replicates (forest = SBM)}} \\
\quad UCI diabetes & 40 & .57 & .59 & \textbf{.72} & .14 [.09, .16] \\
\quad WiDS, encounter-level roles & 50 & .58 & \textbf{.77} & \textbf{.77} & .00 [$-$.02, .02] \\
\quad WiDS, hospital-disjoint roles & 20 & .11 & .24 & \textbf{.25} & .01 [$-$.04, .06] \\
\bottomrule
\end{tabular}

\end{table}

\begin{table}[!hb]
\caption{Sensitivity to \texttt{max\_features} in paired development replicates: observation-level $F_1$ of the
forest pockets (continuous: spectral; binary and real covariates: SBM / Leiden) by the number of candidate
features per split. Default is $p/2$ for continuous outcomes and $\sqrt{p}$ otherwise; --: not run.
Ridge-weighted K-means is shown for reference.}
\label{tab:mtry}
\scriptsize
\begin{tabular}{@{}llrcccc@{}}
\toprule
Design & Scenario & Reps & Default & 1/3 of $p$ & All $p$ & Ridge \\
\midrule
Continuous (forest) & Linear risk & 100 & .59 & -- & .64 & .65 \\
Continuous (forest) & + nonlinear risk & 100 & .43 & -- & .34 & .14 \\
Continuous (forest) & Region on top linear feature & 100 & .64 & -- & .65 & .76 \\
Continuous (forest) & Nonlinear, standardized & 100 & .42 & -- & .33 & .08 \\
Continuous (forest) & + missing values & 100 & .34 & -- & .22 & .11 \\
Binary (SBM / Leiden) & Region on the risk feature & 15 & .83 / .78 & -- & .82 / .79 & .68 \\
Binary (SBM / Leiden) & Nonlinear one-tail region & 15 & .22 / .25 & -- & .43 / .48 & .15 \\
Binary (SBM / Leiden) & Masking & 15 & .19 / .23 & -- & .12 / .17 & .19 \\
Real cov. (SBM / Leiden) & WiDS ventilated, encounter roles & 20 & .62 / .66 & .40 / .49 & .40 / .36 & .77 \\
\bottomrule
\end{tabular}

\end{table}

\phead{Hospital-disjoint versus encounter-level roles.}
The same audits on simulated outcomes for the real patients (main text, Section 5) explain the limited
local resolution in the ICU data. With hospital-disjoint roles, about 70\% of patients carrying a planted
gap sit in pockets that fail the support rule, mostly because each pocket must draw its Black patients
from at least 15 distinct and 10 effective hospitals. Assigning roles by \emph{encounter} while still
treating hospitals as the unit of dependence leaves only 13\% untested and raises the forest pockets'
$F_1$ from 0.11 to 0.58. In a dependence stress check, with outcomes sharing a hospital-level shock and
LightGBM and logistic models fitted to the development rows, the simultaneous intervals of the primary
family (both models, three contrasts and their paired gap) covered every true gap in 100 of 100
replicates in both role designs (Wilson upper bound on non-coverage 0.037). This is one stress design, not a
general proof that encounter-level roles handle hospitals shared between training and audit. Encounter-level roles change the
question, from unseen hospitals to future patients of known hospitals, and are a design choice for the
next audit, not a re-analysis of this one.

\section{Validation of the Local Calibration Error}
\label{app:muval}

Protocol v2 required a validation gate before the new analyses read any audit outcome. Each replicate keeps the real
audit observations, their units, groups and frozen pocket assignments, and draws new outcomes so that the generating
model is calibrated in every pocket (\emph{null}); calibrated in each pocket on average but not at every observation
(\emph{weak null}, ICU only: $p\,[1 + 0.3\,z\,(1-p)]$ with $z = \pm 1$ split at the median heart rate within
each tree-cut pocket, rescaled to the pocket mean; the factor $(1-p)$, added to the protocol's $p(1 + 0.3z)$ in
implementation, bounds the unrescaled perturbation, and final probabilities were range-checked after rescaling); or miscalibrated in a fixed region (\emph{planted error}: true risk 5 points lower
for ventilated stays; 0.5 more conditions for patient-years with at least three prior-year conditions). ICU outcomes
are Bernoulli with a hospital shock $p(1 + v_h(1-p))$, $v_h = s\,U_h$, $U_h \sim \mathrm{Uniform}(-1, 1)$, which keeps
the marginal mean at $p$; $s$ was set from the dispersion of observed-to-expected deaths across non-audit hospitals
(0.68 for LightGBM, 0.56 for APACHE~IVa). Care-management outcomes add residuals resampled within deciles of the fitted
value from development calibration rows and centred within each decile. ICU families hold $\mu$ for both models at
$\alpha = 0.0125$ per map; care-management families hold $\mu$, $\delta$ and $\kappa$ for the three models and two
contrasts at $\alpha = 0.025$ per map ($0.05$ for the combined audit). The false-flag rate counts replicates with any
flagged coordinate whose true value is zero; coverage is the share of replicates in which every tested $\mu$ and
$\delta$ interval covers its true value ($\kappa$ is fixed under this conditional outcome resampling, so these checks do not validate its coverage under repeated sampling of units). Bootstrap $B = 1{,}999$.

\begin{table}[h]
\caption{Validation of the protocol-v2 audits: false-flag rate (\%, 95\% Wilson interval) over true-null coordinates
and simultaneous coverage (\%).}
\label{tab:muval}
\scriptsize
\begin{tabular}{@{}lllrrrlr@{}}
\toprule
Data & Scenario & Generating model & Map & Replicates & Null coords. & False flags & Coverage \\
\midrule
ICU & null & APACHE IVa & SBM & 500 & 3 & 0.2 [0.0, 1.1] & 99.8 \\
ICU & null & APACHE IVa & tree cut & 500 & 4 & 0.2 [0.0, 1.1] & 99.8 \\
ICU & null & LightGBM & SBM & 500 & 3 & 0.0 [0.0, 0.8] & 100.0 \\
ICU & null & LightGBM & tree cut & 500 & 4 & 0.0 [0.0, 0.8] & 100.0 \\
ICU & weak null & APACHE IVa & SBM & 500 & 1 & 0.0 [0.0, 0.8] & 99.8 \\
ICU & weak null & APACHE IVa & tree cut & 500 & 4 & 0.2 [0.0, 1.1] & 99.8 \\
ICU & weak null & LightGBM & SBM & 500 & 1 & 0.2 [0.0, 1.1] & 99.8 \\
ICU & weak null & LightGBM & tree cut & 500 & 4 & 0.4 [0.1, 1.4] & 99.6 \\
ICU & planted error & APACHE IVa & SBM & 500 & 0 & -- & 99.8 \\
ICU & planted error & APACHE IVa & tree cut & 500 & 0 & -- & 100.0 \\
ICU & planted error & LightGBM & SBM & 500 & 0 & -- & 100.0 \\
ICU & planted error & LightGBM & tree cut & 500 & 0 & -- & 100.0 \\
\addlinespace
Care mgmt. & null & cost label & combined & 500 & 36 & 2.2 [1.2, 3.9] & 97.8 \\
Care mgmt. & null & cost label & ridge & 500 & 18 & 1.0 [0.4, 2.3] & 99.0 \\
Care mgmt. & null & cost label & SBM & 500 & 18 & 1.6 [0.8, 3.1] & 98.4 \\
Care mgmt. & null & health label & combined & 500 & 36 & 4.0 [2.6, 6.1] & 96.0 \\
Care mgmt. & null & health label & ridge & 500 & 18 & 1.8 [0.9, 3.4] & 98.2 \\
Care mgmt. & null & health label & SBM & 500 & 18 & 2.4 [1.4, 4.1] & 97.6 \\
Care mgmt. & null & health-mapped score & combined & 500 & 36 & 1.6 [0.8, 3.1] & 98.4 \\
Care mgmt. & null & health-mapped score & ridge & 500 & 18 & 0.6 [0.2, 1.7] & 99.4 \\
Care mgmt. & null & health-mapped score & SBM & 500 & 18 & 1.0 [0.4, 2.3] & 99.0 \\
Care mgmt. & planted error & health label & combined & 500 & 10 & 1.0 [0.4, 2.3] & 98.6 \\
Care mgmt. & planted error & health label & ridge & 500 & 4 & 0.0 [0.0, 0.8] & 99.6 \\
Care mgmt. & planted error & health label & SBM & 500 & 6 & 1.0 [0.4, 2.3] & 98.8 \\
\bottomrule
\end{tabular}
\end{table}

\section{Diabetes Readmission Application}
\label{app:uci}

This application was part of the submitted version (protocol v1, frozen before any audit outcome was read) and is reported here unchanged.

\phead{Diabetes readmission.} The UCI Diabetes 130-US Hospitals data \cite{strack2014diabetes,
clore2014uci} contain 101{,}766 encounters; we exclude 2{,}423 that ended in death or hospice, leaving
99{,}343 encounters from 69{,}990 patients, 11.4\% readmitted within 30 days. We use 43 variables known at
discharge and exclude payer (a socioeconomic proxy) and weight (97\% missing). Patients are the units,
split 40/30/30; the audit has 29{,}782 encounters from 20{,}997 patients. Hospital identifiers are not
released.

\begin{figure}[h]
  \centering
  \includegraphics[width=\linewidth]{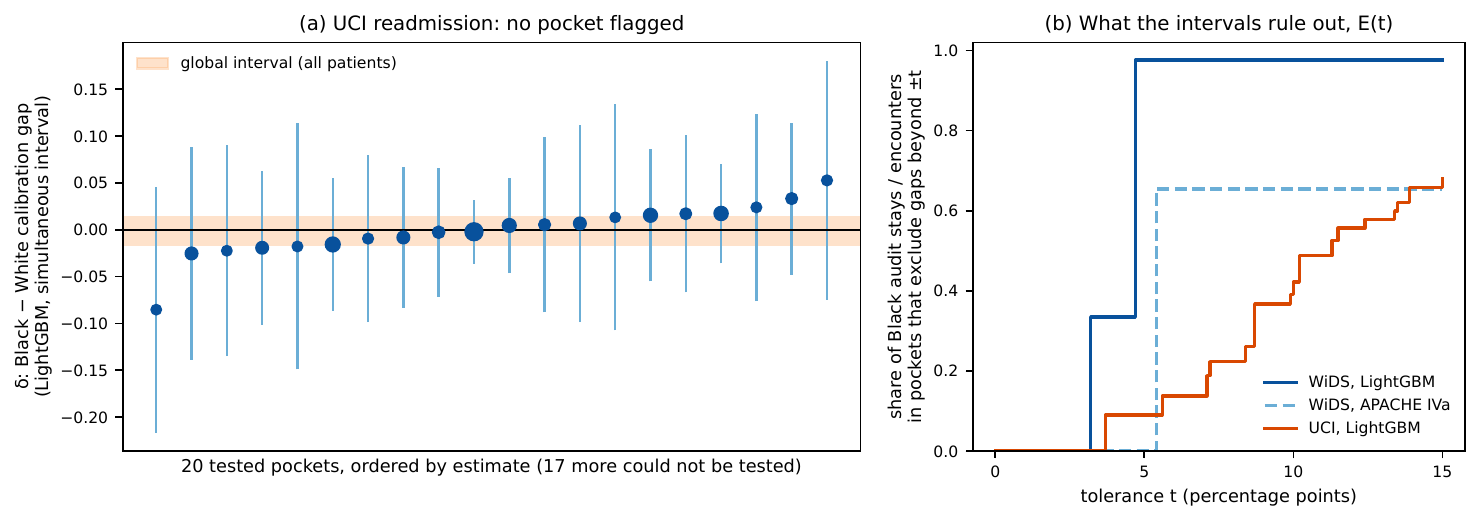}
  \caption{Protocol v1 audits (primary SBM pockets at the full level, Black versus White). (a) Readmission, LightGBM: the 20 testable pockets with simultaneous intervals (marker size proportional to the number of Black encounters); the band is the overall interval. (b) Equivalence share $E(t)$: the share of all Black audit stays (WiDS) or encounters (UCI) whose pocket interval excludes a pocket-average Black--White gap larger than $t$.}
  \label{fig:uci}
  \Description{Left: interval estimates for 20 readmission pockets, all covering zero. Right: step curves of the equivalence share.}
\end{figure}

\phead{Model and results.} LightGBM was fit on development patients and recalibrated with Platt scaling on held-out development patients. Readmission is hard to predict (AUROC 0.66 [0.65, 0.67] on audit encounters), and the overall Black--White calibration gap is near zero ($-0.001$, simultaneous interval $[-0.018, 0.015]$); the calibration error is $+0.003$ [$-0.005$, 0.011] for Black and $+0.005$ [0.000, 0.009] for White encounters (pointwise intervals). The SBM found 37 pockets, of which 20 could be tested for the Black--White contrast, covering 72\% of Black audit encounters (Figure~\ref{fig:uci}a); all intervals cover zero. The finer partition gives weaker bounds than the two coarse ICU pockets: 9\% of Black audit encounters are in pockets whose interval excludes a pocket-average gap above 5 points, and 42\% above 10 points (Figure~\ref{fig:uci}b). Read at the combined audit's per-map level, neither map flags any pocket; the share at 10 points falls to 37\% (forest) and 64\% (ridge). Coarse pockets buy precision; fine pockets buy localization.

\section{Data and Protocols}
\label{app:data}

\phead{Hospital mortality of ICU stays.} The WiDS Datathon 2020 data \cite{lee2020wids}, assembled by MIT's GOSSIS \chg{(Global Open Source Severity of Illness Score)}
initiative \cite{raffa2022gossis} and distributed through PhysioNet \cite{goldberger2000physionet} under its
restricted-access terms (registration and a signed data use agreement; we do not redistribute the data), contain
91{,}713 ICU stays from 147 hospitals; 8.6\% ended in death before hospital discharge. We use about 90
variables recorded in the first 24 hours: demographics and body measurements, admission type, APACHE
physiology components, chronic conditions, and first-day extremes of vital signs and laboratory values.
We exclude first-hour values (they duplicate the first-day extremes), the high-cardinality APACHE
diagnosis code (we keep its body system), and the hospital admission source, which could let the forest
identify hospitals. Each stay also carries the APACHE~IVa predicted probability of death
\cite{zimmerman2006apache}, a deployed clinical score; it is missing or invalid (negative) for 10{,}318
stays, which are excluded from comparisons involving APACHE. Hospitals are the units: 59 for
development, 44 for discovery and 44 for audit (23{,}216 stays), so the audit evaluates hospitals
that neither trained a model nor shaped a pocket.

\phead{Models and protocol.} In WiDS and UCI we fit LightGBM \cite{ke2017lightgbm} and \chgr{a}
\chg{ridge-penalized ($L_2$)} logistic regression on development units and recalibrate both with Platt scaling (a
logistic re-mapping of the predicted probabilities) \cite{platt1999probabilistic} on held-out
development units. The primary contrast is Black versus White
patients; female versus male and Hispanic versus White are secondary. All settings were committed to the repository before any audit outcome was read, and each audit ran
once. That protocol predates the combined audit: it named the SBM forest pockets as the primary map, and
the tree cut of the SBM hierarchy, Leiden, ridge-weighted K-means and the other baselines as pre-declared
comparison maps, each its own family at the full level. Because the bootstrap distribution of every
family was saved, we can also read both maps of the combined audit at its stricter per-map level,
$\alpha/2$, without touching the outcomes again. We report that readout as a supplemental analysis made
after the protocol was frozen; the frozen forest-map audit remains the primary analysis. In WiDS, across the
pre-declared comparison maps (each its own family), the only flag in the primary calibration family was a
borderline LightGBM-versus-APACHE difference between women and men in one plain K-means pocket; with several
separately audited maps, we do not interpret it.

\phead{Care-management data.} The public synthetic data of \citet{obermeyer2019dissecting} (gitlab.com/labsysmed/dissecting-bias,
commit \texttt{f7a3b4d8}, file \texttt{data\_new.csv}, SHA-256 \texttt{5341f90f\ldots4d74}) contain 48{,}784 patient-years (5{,}582 Black,
43{,}202 White) generated with the synthpop package from the authors' analytic file. We found no explicit license
file in that repository; we identify the public release and pinned version and do not redistribute the data. The synthetic rows do not preserve the original patient clustering, so we treat rows as
independent audit units; the intervals do not recover the original patients' sampling uncertainty. The outcome is the number of active chronic conditions in year $t$ (\texttt{gagne\_sum\_t}). The pocket-forming features
are the 148 prior-year (\texttt{\_tm1}) variables: comorbidity indicators and their sum, cost components, laboratory-test counts and range
indicators, medication indicators and age bands; race, sex and every year-$t$ variable are excluded. The assessed models use the same features
plus sex, following the original code; race is excluded from every model. The cost- and health-label models are
lasso regressions (5-fold cross-validated penalty) of $\log(\text{cost}_t + 1)$ and of the number of conditions, fit on
development training rows; these and the commercial score (the health-mapped score) are mapped to the outcome scale by an increasing isotonic
regression fit on development calibration rows. Roles are 40/30/30 by row (seed 20261101), and development rows are
split 75/25 into training and calibration (seed 20261102). Before the protocol was written, we computed the cohort-wide mean number of
conditions by score decile and race, reproducing the published pattern; no pocket-level quantity was computed before the audit.

\phead{Protocols.} Protocol v1 (WiDS, UCI; committed before any audit outcome was read), protocol v2 (care-management data; the
$\mu$ analysis of WiDS, labeled exploratory) and protocol v3 (chest X-rays) were each committed before the corresponding audit, and
every run wrote a manifest recording input hashes, exclusions, environment, configuration and commands. Code and data-access
instructions are at \url{https://github.com/modenesi/pocket-audit}.

\section{Localizing a Local Miscalibration ($\mu$)}
\label{app:musims}

\phead{Design.} The $\delta$ scenarios of the main text plant an error for focal-group observations only. To
evaluate \emph{where}, we re-audited saved replicates with an error planted for \emph{every} observation in a region
$R$: $f = m + s\,\mathbf{1}_R$ for a continuous outcome; for a binary one, positive shifts are capped at
$\max(p_0, 0.995)$, negative shifts are floored at $\min(p_0, 0.005)$, and unshifted predictions remain $p_0$
(masking: $+s$ in one half of the region and $-s$ in the other, so the overall error is near zero). The saved
groupings were held fixed and nothing was rediscovered; only the error-targeted comparator is refit for each
condition. In the continuous ladder (arm P) and in the binary null designs
D0 and DH0, discovery saw only features and outcomes drawn from the truth, so the maps never saw any assessed
model's predictions (the binary level rule projected expected events from the true risk). The binary one-tail and
masking rows reuse the maps of the corresponding $\delta$ designs, whose level rule projected expected events from
that design's predictions; they are audits of maps frozen under the original design (marked $^\ddagger$ in the
tables). Two scenarios move the
region to a feature unrelated to the outcome (continuous $z_7$; binary $x_2$), where outcome-learned maps should do
poorly. Shifts are $s = 0.2$ outcome SD or 5 points (``smaller'', chosen because $\mu$ pools both groups and is
estimated more precisely than $\delta$) and $s = 0.4$ or 10 points (``larger'', as in the $\delta$ designs); both are
reported.

\phead{Audit.} Each map is audited for $\mu$ alone (every pocket plus ``all'') with the settings of the
$\delta$ designs: wild restricted bootstrap with Mammen weights, symmetric max-$t$, $B = 1{,}999$ for continuous and
independent binary designs and $4{,}999$ for the 150-unit binary design, single-map levels $0.05$ (continuous), $0.025$ (binary) and $0.0125$ (binary with 150 units),
the combined audit at half of each, and the support rule on the pooled pocket cell with the new predictions.
A secondary run audits the joint $\mu$ and $\delta$ family of the two primary maps. The group is assigned
independently of everything, so every population $\delta$ is zero; finite-sample truths are the differences between
the group-specific means of the realized shift in each audit pocket, which need not be exactly zero.

\phead{Error-targeted comparator.} The $\mu$ analogue of the disparity-targeted method fits, on graph
observations, a gradient-boosting regression of the assessed model's residual $f(x) - y$ on the features (the same
settings as the disparity-targeted method), and cuts its predictions into as many quantile bins as the forest map
has pockets, with edges from calibration observations. It is refit for every condition. Like slice-discovery
methods, it is organized around one model's errors; our maps are not.

\phead{Scoring.} As in the main text, but over all observations: affected observations (realized bias
$\neq 0$) in pockets flagged in the right direction are true positives, flagged unaffected observations are false
positives, wrong-direction flags count as both, untested and abstained affected observations are misses, and the
combined audit counts conflicting flags as errors. The ``all'' coordinate counts toward false flags, never toward
localization. Coverage is against the finite-population truth, the audit-sample mean of the planted bias in each
pocket. Intervals for differences are paired and pointwise, from a bootstrap of whole replicates.

\phead{Nulls.} Under the \emph{exact} null the assessed predictions equal the truth. Under the \emph{weak} null
each map gets predictions that are wrong within every pocket but whose audit-sample pocket means are exactly zero
($m + 0.4(z - \bar z_c)$, or $p_0 + 0.5\,p_0(1-p_0)(z - w_c)$ with $w_c$ the $p_0(1-p_0)$-weighted mean, where
$z = \pm 1$ by the sign of an outcome-irrelevant feature); it is checked one map at a time.

\phead{Results.} Tables~\ref{tab:mu-matched} and~\ref{tab:mu-same} give every grouping's $F_1$ at the two
shifts, and Table~\ref{tab:mu-errors} the false flags and coverage. At the smaller shift, the combined audit and the
forest map were each favored over every designated alternative in 12 of 15 scenarios (paired pointwise intervals) and
lost to the error-targeted grouping in the two irrelevant-feature scenarios and the binary one-tail region. At the
larger shift, the combined audit beat every alternative in 8, lost to at least one in 6 (adding the standardized
nonlinear design and two masking designs) and was not distinguishable in 1; the forest map beat every alternative in 9.
At the smaller shift, the error-targeted grouping's $F_1$ was 0.04--0.20 in the continuous designs; its residual model
used 2{,}400 graph observations in the $n = 10{,}000$ designs and 4{,}800 in the $n = 20{,}000$ masking designs. In the
masking designs the overall comparison flagged at most 0.7\% of replicates. The joint $\mu$ and $\delta$ family gave
similar $F_1$ (within 0.05 of the $\mu$-only family for the combined audit), and a $\delta$ coordinate was flagged in
at most 6\% of replicates per map; population $\delta$ is zero under independent group assignment, although the
audit-sample truth can differ from zero. With 150 dependent units, false flags were 8.9\% (forest) and 9.1\%
(combined; 41 of 450) at the levels calibrated for $\delta$. Retrospectively, from the same 450 null replicates' smallest
adjusted $p$-values, a single-map level of about 0.0025 gave 4.7\% (21 of 450; Wilson interval 3.1--7.0\%), against
about 0.046 with independent rows (\texttt{calibration\_pglobal.csv}); this has not been confirmed on fresh
replicates, and the dependent-unit localization results come from an audit with inflated false flags.

\begin{table}[h]
\caption{$\mu$ localization at the smaller shift (0.2 outcome SD; 5 points): observation-level $F_1$ pooled over
replicates; bold marks the largest unrounded non-oracle score. Comb.: forest + ridge; K-m.: K-means; Mahal.: Mahalanobis
K-means; Error-t.: error-targeted; --: not available in that batch. $^\ddagger$: maps frozen under the $\delta$
design.}
\label{tab:mu-matched}
\scriptsize
\setlength{\tabcolsep}{3pt}
\begin{tabular}{@{}lrrrrrrrrrrrr@{}}
\toprule
Scenario & Reps & Forest & Comb. & Ridge & Leiden & Risk bins & K-m. & Mahal. & CEM & CEM-3 & Error-t. & Oracle \\
\midrule
Continuous, linear risk & 200 & .67 & \textbf{.73} & .65 & -- & .55 & .17 & -- & -- & -- & .19 & .99 \\
\quad + raw scales & 200 & .68 & \textbf{.74} & .68 & -- & .54 & .17 & -- & -- & -- & .19 & .99 \\
\quad + categorical & 200 & .65 & \textbf{.74} & .68 & -- & .54 & .07 & -- & -- & -- & .18 & .99 \\
\quad + nonlinear (U-shaped region) & 200 & .54 & \textbf{.56} & .33 & -- & .10 & .08 & -- & -- & -- & .18 & .99 \\
Nonlinear; region on top linear feature & 200 & .71 & \textbf{.80} & .78 & -- & .52 & .08 & -- & -- & -- & .20 & .98 \\
Nonlinear; standardized & 200 & \textbf{.55} & .53 & .06 & -- & .10 & .07 & -- & -- & -- & .18 & .99 \\
\quad + missing values & 200 & .42 & \textbf{.44} & .26 & -- & .09 & .07 & -- & -- & -- & .14 & .99 \\
Linear; region on an irrelevant feature & 200 & .08 & .07 & .06 & -- & .06 & .17 & -- & -- & -- & \textbf{.17} & -- \\
Masking, linear ($n = 20{,}000$) & 150 & .52 & \textbf{.59} & .42 & -- & .19 & .02 & -- & -- & .21 & .04 & .99 \\
Masking, nonlinear ($n = 20{,}000$) & 150 & \textbf{.48} & .45 & .05 & -- & .05 & .00 & -- & -- & .22 & .04 & .99 \\
Binary, region on risk feature & 450 & .84 & \textbf{.87} & .80 & .80 & .69 & .71 & .71 & .64 & -- & .69 & -- \\
Binary, region on an irrelevant feature & 450 & .50 & .43 & .21 & .58 & .19 & .74 & .74 & .71 & -- & \textbf{.75} & -- \\
Binary, within-unit dependence & 450 & .40 & \textbf{.41} & .36 & .39 & .30 & .32 & .31 & .28 & -- & .31 & -- \\
Binary, nonlinear one-tail region$^\ddagger$ & 40 & .26 & .28 & .23 & .26 & .22 & .28 & .26 & .00 & -- & \textbf{.69} & -- \\
Binary, masking$^\ddagger$ & 30 & .45 & \textbf{.62} & .56 & .46 & .15 & .01 & .00 & .00 & .15 & .22 & -- \\
\bottomrule
\end{tabular}

\end{table}

\begin{table}[h]
\caption{$\mu$ localization at the larger shift of the $\delta$ designs (0.4 outcome SD; 10 points).}
\label{tab:mu-same}
\scriptsize
\setlength{\tabcolsep}{3pt}
\begin{tabular}{@{}lrrrrrrrrrrrr@{}}
\toprule
Scenario & Reps & Forest & Comb. & Ridge & Leiden & Risk bins & K-m. & Mahal. & CEM & CEM-3 & Error-t. & Oracle \\
\midrule
Continuous, linear risk & 200 & \textbf{.72} & .71 & .72 & -- & .64 & .38 & -- & -- & -- & .56 & .99 \\
\quad + raw scales & 200 & .73 & .72 & \textbf{.74} & -- & .64 & .37 & -- & -- & -- & .56 & .99 \\
\quad + categorical & 200 & .73 & .72 & \textbf{.75} & -- & .64 & .22 & -- & -- & -- & .57 & .99 \\
\quad + nonlinear (U-shaped region) & 200 & \textbf{.60} & .60 & .47 & -- & .30 & .22 & -- & -- & -- & .57 & .99 \\
Nonlinear; region on top linear feature & 200 & .73 & .78 & \textbf{.83} & -- & .67 & .22 & -- & -- & -- & .57 & .98 \\
Nonlinear; standardized & 200 & \textbf{.61} & .51 & .23 & -- & .30 & .22 & -- & -- & -- & .57 & .99 \\
\quad + missing values & 200 & .50 & \textbf{.50} & .39 & -- & .27 & .22 & -- & -- & -- & .48 & .99 \\
Linear; region on an irrelevant feature & 200 & .21 & .23 & .21 & -- & .20 & .40 & -- & -- & -- & \textbf{.57} & -- \\
Masking, linear ($n = 20{,}000$) & 150 & .63 & \textbf{.66} & .56 & -- & .34 & .07 & -- & -- & .60 & .53 & .99 \\
Masking, nonlinear ($n = 20{,}000$) & 150 & .57 & .49 & .20 & -- & .20 & .00 & -- & -- & \textbf{.60} & .53 & .99 \\
Binary, region on risk feature & 450 & .85 & .84 & \textbf{.87} & .79 & .71 & .74 & .74 & .75 & -- & .78 & -- \\
Binary, region on an irrelevant feature & 450 & .51 & .35 & .32 & .55 & .31 & .71 & .71 & .75 & -- & \textbf{.85} & -- \\
Binary, within-unit dependence & 450 & .69 & \textbf{.71} & .67 & .66 & .58 & .59 & .59 & .54 & -- & .66 & -- \\
Binary, nonlinear one-tail region$^\ddagger$ & 40 & .34 & .34 & .33 & .35 & .32 & .38 & .37 & .00 & -- & \textbf{.86} & -- \\
Binary, masking$^\ddagger$ & 30 & .63 & .68 & .60 & .62 & .33 & .04 & .03 & .00 & .55 & \textbf{.73} & -- \\
\bottomrule
\end{tabular}

\end{table}

\begin{table}[h]
\caption{False flags of the $\mu$ family (share of replicates with any flagged coordinate, \%) under the exact null
(predictions equal the truth) and the weak null (errors that cancel in every pocket, one map at a time), and
simultaneous coverage of the audit-sample pocket errors at the smaller shift, among replicates with evaluable
coverage. Single-map levels: 0.05 (continuous), 0.025 (binary), 0.0125 (binary, 150 units); the combined audit runs
each map at half. Others (max): the largest rate among the other groupings. ``Reps'' counts exact-null runs, except that the continuous irrelevant-feature row lists its 200 coverage replicates;
continuous-ladder coverage uses 200 replicates; in the 150-unit design combined coverage is 396 of 407, and 43
replicates lack evaluable coverage in both maps. Irrelevant-feature rows reuse the null runs of the maps above them.}
\label{tab:mu-errors}
\scriptsize
\setlength{\tabcolsep}{3pt}
\begin{tabular}{@{}lrrrrrrrr@{}}
\toprule
 & \multicolumn{4}{c}{Exact null: false flags, \%} & \multicolumn{2}{c}{Weak null, \%} & \multicolumn{2}{c}{Coverage, \%} \\
\cmidrule(lr){2-5}\cmidrule(lr){6-7}\cmidrule(lr){8-9}
Scenario & Reps & Forest & Comb. & Others (max) & Forest & Ridge & Forest & Comb. \\
\midrule
Continuous, linear risk & 500 & 4.8 & 3.6 & 6.0 & 2.0 & 1.6 & 94.5 & 94.5 \\
\quad + raw scales & 500 & 4.6 & 3.8 & 5.8 & 2.0 & 1.4 & 94.0 & 95.0 \\
\quad + categorical & 500 & 4.6 & 3.4 & 7.2 & 1.0 & 0.4 & 94.5 & 96.5 \\
\quad + nonlinear (U-shaped region) & 500 & 3.6 & 3.8 & 5.6 & 1.0 & 1.2 & 96.5 & 96.0 \\
Nonlinear; region on top linear feature & 500 & 3.6 & 3.8 & 5.6 & 1.0 & 1.2 & 96.0 & 95.5 \\
Nonlinear; standardized & 500 & 4.6 & 4.6 & 5.6 & 1.4 & 1.2 & 95.5 & 97.5 \\
\quad + missing values & 500 & 6.0 & 4.4 & 4.8 & 1.0 & 1.8 & 93.0 & 96.5 \\
Linear; region on an irrelevant feature & 200 & -- & -- & -- & -- & -- & 94.5 & 95.0 \\
Masking, linear ($n = 20{,}000$) & 150 & 6.0 & 4.7 & 5.3 & 2.0 & 2.0 & 94.7 & 95.3 \\
Masking, nonlinear ($n = 20{,}000$) & 150 & 5.3 & 5.3 & 4.7 & 3.3 & 2.0 & 94.7 & 94.0 \\
Binary, region on risk feature & 450 & 2.9 & 3.1 & 4.0 & 2.9 & 2.7 & 97.3 & 97.1 \\
Binary, region on an irrelevant feature & 450 & -- & -- & -- & -- & -- & 96.9 & 97.1 \\
Binary, within-unit dependence & 450 & 8.9 & 9.1 & 10.2 & 9.1 & 10.7 & 96.3 & 97.3 \\
Binary, nonlinear one-tail region$^\ddagger$ & 40 & 5.0 & 0.0 & 10.0 & -- & -- & 97.5 & 97.5 \\
Binary, masking$^\ddagger$ & 30 & 3.3 & 3.3 & 3.3 & -- & -- & 96.7 & 96.7 \\
\bottomrule
\end{tabular}

\end{table}

\section{A Chest X-Ray Model on NIH ChestX-ray14}
\label{app:cxr}

\phead{Protocol.} This application was added for the resubmission under a protocol frozen and committed
before any label was read (protocol v3). Before that point, the labels were parsed only by a smoke test that replaced
them with random ones. It audits agreement with the NIH \emph{Effusion} label, which is mined from radiology reports
by natural-language processing, not clinical truth \cite{oakdenrayner2020exploring}. The data are the 112{,}120
frontal images of 30{,}805 patients in ChestX-ray14 \cite{wang2017chestxray}. We audit each patient's first image
(lowest follow-up number), so every image is a different patient; follow-up is not a pocket-forming feature. A smoke
test with random labels had shown $\mu$ intervals that were too narrow under strong repeated-image dependence (up to
184 images per patient), as in the 150-unit simulations of Section~\ref{app:musims}. Roles are 40/30/30 by patient.

\phead{Models.} The assessed models are two pretrained TorchXRayVision DenseNet-121 classifiers
\cite{cohen2022xrv}, never trained on NIH images: one trained on CheXpert and one on MIMIC-CXR. Their Effusion logits
are mapped to probabilities by increasing isotonic regressions fit on all 12{,}322 development images (Effusion
prevalence 3.97\%), with probabilities clipped to $[0.001, 0.999]$, so the audited objects are recalibrated
pipelines, and $\kappa$ compares the two pipelines. On development images, CheXpert's raw mean prediction was 16.1\%
and MIMIC's 10.6\%; after remapping both were 3.97\%. Development AUC was 0.812 before and 0.817 after remapping for
CheXpert (36 isotonic steps) and 0.803 and 0.809 for MIMIC (32 steps); the remapped values are on the images used to
fit the maps.

\phead{Pockets and inference.} Pockets are learned from age, view position (AP or PA) and 32 principal
components (fit on discovery images) of the penultimate features of a third classifier, trained only on PadChest.
Patient sex is excluded because it is the audited contrast; chest images encode sex, so the similarity is not
sex-free, and we report each pocket's sex composition. The forest map selected hierarchy level 2 with 17 pockets; the
ridge map also has 17. Each map is one family holding both models' $\mu$ and female-minus-male $\delta$ and the
CheXpert-minus-MIMIC paired gap $\kappa$, at $\alpha/2 = 0.025$. Each compared cell needs at least 100 images, 10
expected events and non-events from each relevant model's predictions, 30 distinct and 20 effective patients, and no
patient above 10\% (with one image per patient, the patient counts equal the image counts). Inference uses 4{,}999
restricted Mammen bootstrap draws, symmetric max-$t$ intervals and the binary standard-error floor. Tolerances are
$\pm 2$ and $\pm 5$ points; Brier-loss contrasts are not reported.

\phead{Validation gate.} Before the audit, conditional simulations on the audit images, their sexes and frozen
pockets (300 replicates per scenario, $B = 4{,}999$) passed the protocol's gate (Table~\ref{tab:cxr-gate}). Under the
independent generating-model nulls, the combined audit falsely flagged at least one true-null coordinate in
0.7--1.3\% of replicates; under the weak null, which makes the generating model's pocket-average $\mu$ zero without
making every sex gap zero, each map flagged at most 1.0\%. Coverage was assessed for the tested $\mu$ and $\delta$
coordinates, including the other model's nonzero ones, and was 97.3--99.3\%. In the planted-effect scenario (AP
over-prediction of up to 5 points, clipped), the generating model's non-null $\mu$ coordinates had mean absolute
truth 0.85--0.93 points and mean interval half-width 3.2--3.8 points, with 2.0\% (ridge) and 4.4\% (forest) of these tested coordinates flagged;
the gate assesses error control and coverage, not power. These 300-replicate checks are consistent with control for
the evaluated design; they do not validate $\kappa$ coverage under repeated sampling of patients.

\phead{Audit.} The audit sample holds 9{,}241 first images; 3.9\% carry the Effusion label, and both recalibrated
models predict 4.0\% on average, so neither shows a detectable overall error (CheXpert $+0.05$ points
$[-0.58, 0.67]$; MIMIC $+0.09$ $[-0.53, 0.72]$). Locally they err in both directions (Table~\ref{tab:cxr-audit}). In
forest pocket 4, the discovery images had an Effusion prevalence of 50.2\% against mean predictions of 15.8\%
(CheXpert) and 20.2\% (MIMIC); on its 257 independent audit images (2.8\%), the mean errors were $-27.6$ and $-23.0$
points. Ridge pocket 15, which shares 185 of its 229 audit images with forest pocket 4, also showed under-prediction
($-30.9$ and $-25.5$). Both models over-predicted in forest pocket 16 (4.1\% of audit images; 95\% of its discovery
images were AP): $+8.0$ $[2.3, 13.7]$ and $+5.4$ $[0.1, 10.7]$ points. Smaller over-predictions were flagged in the
largest, low-prevalence pocket ($+1.3$ and $+1.4$). Table~\ref{tab:cxr-combined} gives the share of each sex's
images that inherit an over-, under- or conflicting flag in the combined audit.

No female--male gap was flagged. Both models' overall gaps are bounded within $\pm 2$ points (CheXpert $-0.41$
$[-1.67, 0.85]$; MIMIC $-0.72$ $[-1.97, 0.54]$), but local sex-gap tests were unavailable for 53\% of women's and 41\%
of men's images in the forest map, and 69\% and 72\% in the ridge map (Table~\ref{tab:cxr-support}; untested shares
include abstentions); pocket-average gaps beyond 5 points are ruled out for 35\% of women's images in the forest map.
In the under-prediction pocket the estimated gap was $-12$ points (women under-predicted more), with an interval of
$[-31, 8]$. The only paired-gap flag was a 1.0-point difference between the two models in ridge pocket 5.
Table~\ref{tab:cxr-counts} gives each pocket's audit counts by sex and its discovery prevalence.

\phead{Reading.} After development-sample recalibration, both pipelines showed small overall errors but
substantial local errors of opposite signs against the report-derived label, including over-prediction in an
AP-rich pocket. Whether a further global or pocket-specific recalibration would remove them needs evaluation on
independent data. The sex comparison illustrates the limits of support with a rare label.

\begin{table}[h]
\caption{Chest X-ray validation gate: share of replicates with any falsely flagged true-null coordinate (95\% Wilson
interval) and simultaneous coverage of the tested $\mu$ and $\delta$ coordinates. The combined audit runs each map at
$\alpha/2 = 0.025$. The planted-effect scenario has no true-null coordinates.}
\label{tab:cxr-gate}
\scriptsize
\begin{tabular}{@{}llrrrr@{}}
\toprule
Scenario & Generating model & Map & False flags, \% [Wilson] & Coverage, \% & Reps \\
\midrule
AP over-prediction (up to 5 points) & CheXpert & ridge & -- (no true-null coordinates) & 97.7 & 300 \\
AP over-prediction (up to 5 points) & CheXpert & forest & -- (no true-null coordinates) & 99.0 & 300 \\
AP over-prediction (up to 5 points) & CheXpert & combined & -- (no true-null coordinates) & 97.3 & 300 \\
Null & CheXpert & ridge & 0.3 [0.1, 1.9] & 99.3 & 300 \\
Null & CheXpert & forest & 0.3 [0.1, 1.9] & 99.0 & 300 \\
Null & CheXpert & combined & 0.7 [0.2, 2.4] & 98.3 & 300 \\
Null & MIMIC & ridge & 0.7 [0.2, 2.4] & 99.3 & 300 \\
Null & MIMIC & forest & 0.7 [0.2, 2.4] & 99.3 & 300 \\
Null & MIMIC & combined & 1.3 [0.5, 3.4] & 98.7 & 300 \\
Weak null & CheXpert & ridge & 1.0 [0.3, 2.9] & 98.7 & 300 \\
Weak null & CheXpert & forest & 0.0 [0.0, 1.3] & 99.3 & 300 \\
\bottomrule
\end{tabular}

\end{table}

\begin{table*}[h]
\caption{Chest X-ray audit, first image per patient: $\mu$ (predicted minus observed Effusion rate), the female--male
gap $\delta$ and the paired gap $\kappa$ (CheXpert minus MIMIC), in percentage points, with simultaneous intervals
(each map at $\alpha/2 = 0.025$; $^*$ flagged). Images: audit images per pocket; Female, Age and AP: discovery-image
profile of the pocket; untested: fails the support rule. Pocket numbers are arbitrary labels.}
\label{tab:cxr-audit}
\scriptsize
\setlength{\tabcolsep}{2.5pt}
\begin{tabular}{@{}llrrrrrrrrr@{}}
\toprule
Map & Pocket & Images & Female & Age & AP & $\mu$ CheXpert & $\mu$ MIMIC & $\delta$ CheXpert & $\delta$ MIMIC & $\kappa$ \\
\midrule
Forest & all & 9,241 & -- & -- & -- & +0.05 [-0.58, +0.67] & +0.09 [-0.53, +0.72] & -0.41 [-1.67, +0.85] & -0.72 [-1.97, +0.54] & +0.31 [-0.00, +0.62] \\
 & 0 & 295 & 47\% & 39 & 95\% & +2.1 [-1.4, +5.7] & +3.3 [-0.7, +7.4] & untested & untested & untested \\
 & 1 & 388 & 44\% & 53 & 4\% & +5.0 [+0.5, +9.4]$^*$ & +2.3 [-1.4, +6.0] & +1.2 [-5.0, +7.4] & -0.0 [-5.9, +5.9] & +1.2 [-0.3, +2.7] \\
 & 2 & 133 & 50\% & 60 & 1\% & untested & untested & untested & untested & untested \\
 & 3 & 2,074 & 45\% & 46 & 3\% & +1.3 [+0.4, +2.3]$^*$ & +1.4 [+0.4, +2.4]$^*$ & -0.7 [-2.2, +0.7] & -1.0 [-2.5, +0.5] & +0.3 [-0.0, +0.5] \\
 & 4 & 257 & 45\% & 54 & 27\% & -27.6 [-37.5, -17.7]$^*$ & -23.0 [-33.0, -13.0]$^*$ & -11.7 [-31.4, +8.0] & -12.1 [-32.0, +7.7] & +0.4 [-5.1, +5.9] \\
 & 5 & 253 & 55\% & 45 & 17\% & untested & -2.9 [-8.0, +2.3] & untested & untested & untested \\
 & 6 & 99 & 22\% & 48 & 72\% & untested & untested & untested & untested & untested \\
 & 7 & 58 & 30\% & 37 & 12\% & untested & untested & untested & untested & untested \\
 & 8 & 204 & 44\% & 47 & 5\% & -8.4 [-16.9, +0.1] & -9.1 [-17.8, -0.4]$^*$ & untested & untested & untested \\
 & 9 & 154 & 63\% & 52 & 28\% & -1.7 [-12.4, +9.1] & -1.7 [-12.6, +9.2] & untested & untested & untested \\
 & 10 & 54 & 45\% & 40 & 3\% & untested & untested & untested & untested & untested \\
 & 11 & 577 & 74\% & 47 & 1\% & +1.0 [-1.4, +3.3] & +0.5 [-1.7, +2.7] & untested & untested & untested \\
 & 12 & 550 & 77\% & 37 & 1\% & untested & +1.6 [-0.8, +4.0] & untested & untested & untested \\
 & 13 & 1,318 & 52\% & 30 & 2\% & +1.0 [-0.1, +2.1] & +1.0 [-0.1, +2.1] & untested & untested & untested \\
 & 14 & 1,812 & 30\% & 54 & 1\% & +0.0 [-1.1, +1.1] & -0.2 [-1.3, +0.9] & -0.3 [-2.8, +2.2] & -0.4 [-2.9, +2.2] & +0.0 [-0.3, +0.4] \\
 & 15 & 490 & 19\% & 48 & 4\% & +1.1 [-1.1, +3.4] & +2.0 [-0.6, +4.6] & untested & untested & untested \\
 & 16 & 376 & 48\% & 43 & 95\% & +8.0 [+2.3, +13.7]$^*$ & +5.4 [+0.1, +10.7]$^*$ & +3.5 [-5.3, +12.2] & +1.5 [-7.6, +10.6] & +1.9 [-1.6, +5.5] \\
\midrule
Ridge & all & 9,241 & -- & -- & -- & +0.05 [-0.59, +0.68] & +0.09 [-0.54, +0.73] & -0.41 [-1.69, +0.86] & -0.72 [-1.99, +0.55] & +0.31 [-0.01, +0.62] \\
 & 0 & 553 & 67\% & 38 & 1\% & untested & +1.5 [-0.5, +3.4] & untested & untested & untested \\
 & 1 & 866 & 50\% & 53 & 2\% & +0.3 [-1.2, +1.9] & +0.0 [-1.4, +1.4] & untested & untested & untested \\
 & 2 & 64 & 49\% & 7 & 55\% & untested & untested & untested & untested & untested \\
 & 3 & 286 & 33\% & 51 & 71\% & +0.1 [-5.8, +6.0] & +0.7 [-5.3, +6.7] & untested & untested & untested \\
 & 4 & 717 & 42\% & 39 & 4\% & +1.2 [-0.5, +2.8] & +1.4 [-0.3, +3.2] & untested & untested & untested \\
 & 5 & 873 & 47\% & 56 & 5\% & +2.6 [+0.3, +4.9]$^*$ & +1.4 [-0.6, +3.4] & -0.5 [-3.9, +2.9] & -1.5 [-4.8, +1.8] & +1.0 [+0.3, +1.7]$^*$ \\
 & 6 & 547 & 46\% & 43 & 1\% & untested & untested & untested & untested & untested \\
 & 7 & 376 & 72\% & 35 & 0\% & untested & untested & untested & untested & untested \\
 & 8 & 306 & 54\% & 46 & 61\% & +2.0 [-5.7, +9.7] & -1.6 [-9.3, +6.2] & +1.9 [-13.7, +17.5] & -0.4 [-16.1, +15.3] & +2.3 [-2.4, +7.1] \\
 & 9 & 865 & 34\% & 51 & 1\% & +1.1 [-0.1, +2.3] & +1.5 [+0.1, +2.8]$^*$ & untested & untested & untested \\
 & 10 & 556 & 53\% & 52 & 24\% & -0.5 [-4.4, +3.4] & -1.7 [-5.6, +2.2] & +3.8 [-4.0, +11.7] & +3.8 [-4.1, +11.6] & +0.0 [-1.5, +1.6] \\
 & 11 & 810 & 27\% & 54 & 3\% & +1.6 [-0.3, +3.5] & +1.5 [-0.4, +3.4] & untested & untested & untested \\
 & 12 & 601 & 29\% & 53 & 21\% & -2.4 [-5.9, +1.0] & -2.5 [-5.9, +1.0] & untested & untested & untested \\
 & 13 & 249 & 57\% & 24 & 32\% & +3.8 [-0.8, +8.5] & untested & untested & untested & untested \\
 & 14 & 568 & 51\% & 30 & 5\% & untested & untested & untested & untested & untested \\
 & 15 & 229 & 51\% & 54 & 12\% & -30.9 [-41.6, -20.2]$^*$ & -25.5 [-36.4, -14.6]$^*$ & -12.1 [-33.4, +9.2] & -12.6 [-34.3, +9.1] & +0.5 [-5.5, +6.6] \\
 & 16 & 775 & 47\% & 36 & 12\% & +1.2 [-0.7, +3.1] & +1.1 [-0.8, +3.1] & +0.3 [-2.8, +3.4] & -0.2 [-3.3, +2.8] & +0.5 [-0.1, +1.2] \\

\bottomrule
\end{tabular}

\end{table*}

\begin{table}[h]
\caption{Chest X-ray support by sex: shares of all audit images of each sex that abstained or fell in pockets
untested for each coordinate (untested shares include abstentions; the $\delta$ shares apply to both models and
equal the $\kappa$ shares).}
\label{tab:cxr-support}
\scriptsize
\begin{tabular}{@{}llrr@{}}
\toprule
Map & Share of images & Female, \% & Male, \% \\
\midrule
Forest & Abstained & 1.9 & 1.4 \\
 & Untested $\mu$, CheXpert & 18.1 & 10.4 \\
 & Untested $\mu$, MIMIC & 5.3 & 5.3 \\
 & Untested $\delta$ and $\kappa$ & 53.3 & 41.2 \\
Ridge & Abstained & 0.0 & 0.0 \\
 & Untested $\mu$, CheXpert & 29.4 & 16.9 \\
 & Untested $\mu$, MIMIC & 25.0 & 14.6 \\
 & Untested $\delta$ and $\kappa$ & 69.0 & 71.6 \\
\bottomrule
\end{tabular}

\end{table}

\begin{table}[h]
\caption{Chest X-ray combined audit: shares of all audit images of each sex inheriting an over- or under-prediction
flag for $\mu$ from either map, or conflicting flags (reported separately and excluded from the other two).}
\label{tab:cxr-combined}
\scriptsize
\begin{tabular}{@{}llrr@{}}
\toprule
Model & Flag inherited & Female, \% & Male, \% \\
\midrule
CheXpert & over & 36.4 & 37.0 \\
 & under & 3.4 & 3.1 \\
 & conflict & 0.02 & 0.06 \\
MIMIC & over & 29.9 & 33.3 \\
 & under & 5.3 & 5.5 \\
 & conflict & 0.02 & 0.04 \\
\bottomrule
\end{tabular}

\end{table}

\begin{table}[h]
\caption{Chest X-ray pockets: audit images by sex (one image per patient, so distinct and effective patients equal
the counts and the largest patient share is $1/n$) and the Effusion prevalence of the pocket's discovery images.
Forest abstentions: 81 female and 68 male audit images (discovery prevalence 25.6\%).}
\label{tab:cxr-counts}
\scriptsize
\begin{tabular}{@{}lrrrrrr@{}}
\toprule
 & \multicolumn{3}{c}{Forest} & \multicolumn{3}{c}{Ridge} \\
\cmidrule(lr){2-4}\cmidrule(lr){5-7}
Pocket & Female & Male & Prev., \% & Female & Male & Prev., \% \\
\midrule
all & 4,357 & 4,884 & -- & 4,357 & 4,884 & -- \\
0 & 132 & 163 & 2.0 & 335 & 218 & 0.7 \\
1 & 202 & 186 & 3.4 & 417 & 449 & 1.2 \\
2 & 67 & 66 & 5.6 & 25 & 39 & 0.0 \\
3 & 981 & 1,093 & 0.5 & 91 & 195 & 10.3 \\
4 & 124 & 133 & 50.2 & 351 & 366 & 1.2 \\
5 & 135 & 118 & 5.2 & 394 & 479 & 2.0 \\
6 & 21 & 78 & 5.1 & 292 & 255 & 0.6 \\
7 & 34 & 24 & 3.5 & 304 & 72 & 0.5 \\
8 & 83 & 121 & 18.6 & 165 & 141 & 21.8 \\
9 & 85 & 69 & 30.3 & 309 & 556 & 0.8 \\
10 & 30 & 24 & 1.7 & 292 & 264 & 8.0 \\
11 & 401 & 176 & 1.3 & 218 & 592 & 1.4 \\
12 & 420 & 130 & 1.4 & 195 & 406 & 5.8 \\
13 & 740 & 578 & 0.6 & 144 & 105 & 0.8 \\
14 & 562 & 1,250 & 0.8 & 325 & 243 & 0.5 \\
15 & 92 & 398 & 1.2 & 118 & 111 & 54.7 \\
16 & 167 & 209 & 6.2 & 382 & 393 & 2.3 \\
\bottomrule
\end{tabular}

\end{table}

\end{document}